\documentclass[journal,twoside,web]{ieeecolor}
\usepackage{jsen}
\usepackage{cite}
\usepackage{amsmath,amssymb,amsfonts}
\usepackage{algorithmic}
\usepackage{graphicx}
\usepackage{textcomp}
\usepackage{wrapfig}
\usepackage{array,multirow}
\usepackage{setspace}
\usepackage{amsmath,amsfonts,amssymb,booktabs}
\usepackage{graphicx}
\usepackage{epstopdf}
\usepackage{soul}
\usepackage{color, xcolor}
\usepackage{color}
\usepackage{multirow}
\usepackage{subfigure}
\def\BibTeX{{\rm B\kern-.05em{\sc i\kern-.025em b}\kern-.08em
    T\kern-.1667em\lower.7ex\hbox{E}\kern-.125emX}}
\definecolor{abstractbg}{rgb}{0.89804,0.94510,0.83137}
\begin{document}
\title{Spatial-information Guided Adaptive Context-aware Network for Efficient RGB-D Semantic Segmentation}
\author{Yang Zhang, Chenyun Xiong, Junjie Liu, Xuhui Ye, and Guodong Sun
\thanks{\emph{(Corresponding author: Guodong Sun.)}}
\thanks{Y. Zhang, C. Xiong, J. Liu, X. Ye, and G. Sun are with the School of Mechanical Engineering, 
	Hubei University of Technology, Wuhan 430068, China (e-mail: yzhangcst@hbut.edu.cn; cyx@hbut.edu.cn; 102210139@hbut.edu.cn; yxh89@hbut.edu.cn; sunguodong@hbut.edu.cn).}
\thanks{Y. Zhang is also with the Department of Electronic Engineering, The Chinese University of Hong Kong, Hong Kong; and with the National Key Laboratory for Novel Software Technology, Department of Computer Science and Technology, Nanjing University, Nanjing 210023, China.}}
\IEEEtitleabstractindextext{%
\fcolorbox{abstractbg}{abstractbg}{%
\begin{minipage}{\textwidth}%
\begin{wrapfigure}[18]{r}{3.0 in}%
\includegraphics[width=2.9 in]{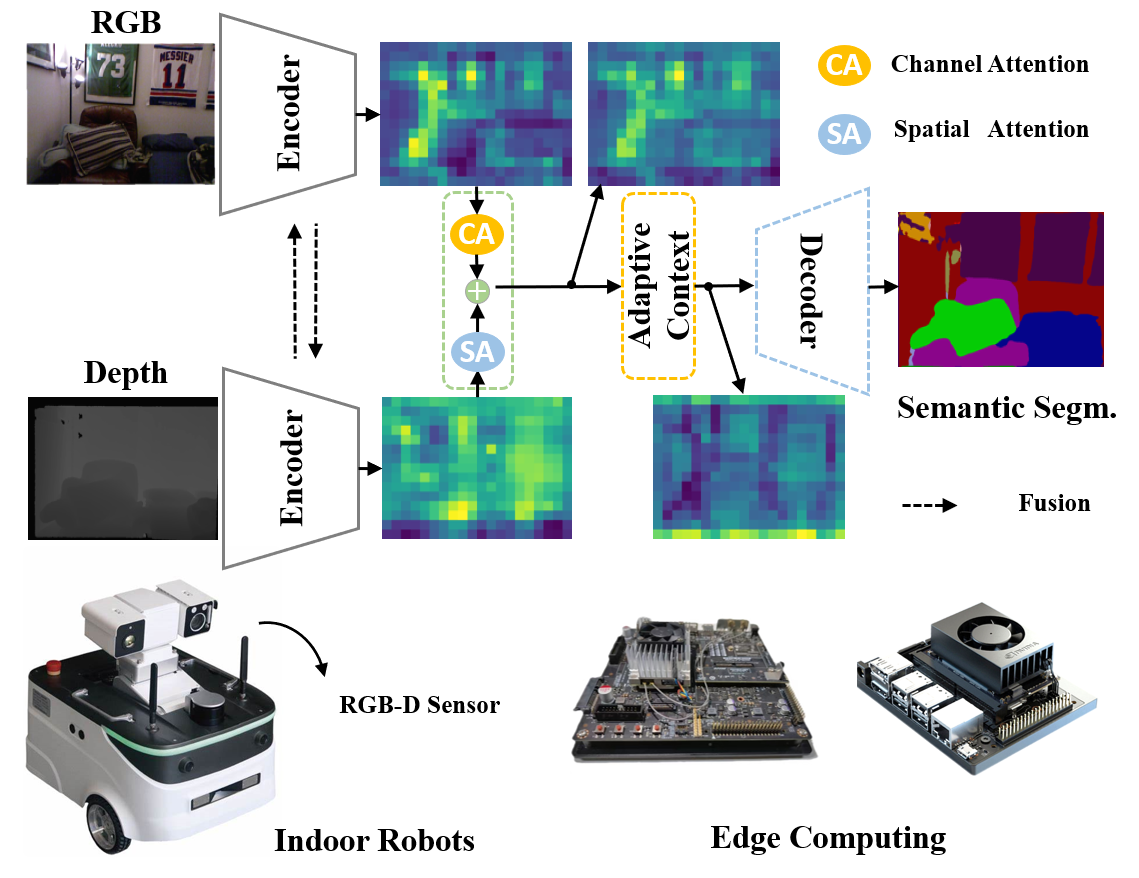}%
\end{wrapfigure}%
\begin{abstract}
Efficient RGB-D semantic segmentation has received considerable attention in mobile robots, which plays a vital role in analyzing and recognizing environmental information. According to previous studies, depth information can provide corresponding geometric relationships for objects and scenes, but actual depth data usually exist as noise. To avoid unfavorable effects on segmentation accuracy and computation, it is necessary to design an efficient framework to leverage cross-modal correlations and complementary cues. In this paper, we propose an efficient lightweight encoder-decoder network that reduces the computational parameters and guarantees the robustness of the algorithm. Working with channel and spatial fusion attention modules, our network effectively captures multi-level RGB-D features. A globally guided local affinity context module is proposed to obtain sufficient high-level context information. The decoder utilizes a lightweight residual unit that combines short- and long-distance information with a few redundant computations. Experimental results on NYUv2, SUN RGB-D, and Cityscapes datasets show that our method achieves a better trade-off among segmentation accuracy, inference time, and parameters than the state-of-the-art methods. 
\end{abstract}

\begin{IEEEkeywords}
RGB-D semantic segmentation, spatial and channel attention, encoder-decoder framework, efficient 
\end{IEEEkeywords}
\end{minipage}}}

\maketitle

\section{Introduction}
\label{sec:introduction}
\IEEEPARstart{S}{emantic} segmentation aims to classify objects in a scene at the pixel level, which helps us understand the message conveyed by the scene and predict the behavior of the target with an RGB-D sensor. This technique is currently widely used in edge computing applications such as autonomous driving, medical image analysis, robotics, and other intelligent fields~\cite {9940618}. 
However, there are still opportunities and challenges for RGB-D semantic segmentation. Multi-modal information cannot be discarded because depth image contains rich spatial information available to determine the location of objects. Unfortunately, due to the interference of cameras and the external environment, we need an efficient mechanism to ignore other insignificant factors (\textit{e.g.} noise) from depth inputs. Moreover, it is well known that the spatial information in the depth map varies from semantic information in RGB images, so different attention mechanisms should be applied~\cite{Gated_Fusion}. Squeeze-and-excitation~\cite{SENet} is considered one of the representative methods for channel dimension. And spatial attention matching with channel attention has the ability to achieve an optimal solution.

From the perspective of encoder-decoder frameworks~\cite{RAFNet}, the encoder uses downsampling for feature extraction, and the decoder uses upsampling to restore the size of the feature map. Summation-based skip connection~\cite{EFCN} connects the information before and after the network, which can effectively reduce gradient disappearance and network degradation problems, reducing the number of layers. The attention-based network~\cite{CMANet} can capture global dependencies and long-range contextual information. Nevertheless, these methods still require a complex backbone to get rich semantic information consuming large computing resources~\cite{LF-YOLO}. Based on such approaches, some methods~\cite{TSNet} build an adaptive context module to enhance the features of the encoder, which is the information bridge between the encoder and decoder. However, it is still difficult to achieve both efficient and accurate segmentation. Early existing methods~\cite{ICNet} focus on reducing the size of input images. Although they can reach fast inference speed for less computation, these methods lead to some problems, such as information loss and unclear target edges. Some special structures reduce computation by decreasing the number of channels and convolutional layers~\cite{min12050526}. 
Other lightweight and efficient structures are used as substitutes for general convolutions, such as depthwise separable convolutions, which are computationally efficient while low accuracy. Moreover, Atrous convolution~\cite{DCNV2} can expand the receptive field and reduce the loss of resolution caused by downsampling. 

According to the above analysis, we balance accuracy, speed, and parameters. Therefore, we design a spatial information-guided adaptive context-aware network (SGACNet\footnote{The code will be available at https://github.com/MVME-HBUT/SGACNet}) for efficient RGB-D semantic segmentation. For different depth and RGB inputs, the proposed attention fusion module enhances their respective features and then takes fused channel-spatial information to guide the next encoder layer. We further propose a multi-scale context module to capture rich context information adaptively. In particular, a fine-designed lightweight residual unit is introduced to help reduce the amount of parameter calculation. Experimental results on NYUv2, SUN RGB-D, and Cityscapes datasets show that our method implements efficient semantic segmentation with higher accuracy for indoor scenes and outdoor urban landscapes. Our method achieves competitive performance with fewer model parameters for resource-constrained scenarios than state-of-the-art approaches.

The main contributions of this article are as follows.

$\bullet $ We explore the correlations and complementary cues between RGB and depth images through a novel light-weighted encoder-decoder framework for the improvement of RGB-D semantic segmentation performance.

$\bullet $ We propose dual-branch attention fusion and adaptive pyramid context modules to learn more robust deep representations and rich cross-modal information efficiently. 

$\bullet $ The proposed method achieves competitive performance with fewer model parameters against the state-of-the-art methods on three public benchmark datasets.

The remainder of this paper is structured as follows. We will introduce the related work in Section~\ref{sec:Related Work}. Section~\ref{sec:Proposed Method} describes our innovative methods in detail. To explore the factors that influence final indicators, we then conduct ablation experiments in Section~\ref{sec:Experiments}, and validate the best methods on three datasets. The final conclusion is summarized in Section~\ref{sec:Conclusions}.

\section{Related Work}\label{sec:Related Work}
More recently, an encoder-decoder framework has become one of the classical structures for RGB-D semantic segmentation. Visual robots need to focus on understanding useful semantic information, which is adapted to different environments efficiently. 
\subsection{RGB-D Semantic Segmentation}
Semantic segmentation is a computer vision task that classifies objects in the scene based on their pixel-level content, typically using RGB images as input~\cite{DRD, DSNet, FDNet, MTI-Net, RefineNet,DCNV2}. However, lighting conditions easily affect RGB images, which may cause segmentation errors. In some fields, like medical image segmentation and indoor scene exploration, RGB images may not be sufficient to meet the requirements, and other multi-modal images are required for analysis.

Recently, depth information has become increasingly popular in semantic segmentation~\cite{SGNet, LDFNet,3DGNN, ESANet, MCN-DRM, ACNet, VCD, LSTMCF}. Yan \textit{et al.}~\cite{RAFNet} incorporated attention mechanisms to enhance image recognition accuracy instead of simply linking RGB and depth information. Chen \textit{et al.}~\cite{ SGNet } employed inter-group augmentation modules to enhance feature representation and discrimination, particularly in tasks that demand accurate spatial context understanding from depth maps. Based on luminance information, Hung \textit{et al.}~\cite{LDFNet} applied depth information to distinguish the contour features. Cao \textit{et al.}~\cite{3DGNN } excelled in processing 3D data, such as point clouds or grids, and extracting features with graph convolutional layers. Seichter \textit{et al.}~\cite{ESANet} mitigated the impact of depth noise by employing feature fusion and attention mechanisms. To enhance the performance of the semantic segmentation with dual input, Zheng \textit{et al.}~\cite{ MCN-DRM} and Hu \textit{et al.}~\cite{ ACNet} shared the common approach of utilizing multi-scale context information. However, the segmentation methods mentioned above, also including TSNet~\cite{ TSNet }, VCD~\cite{VCD}, and LSTMCF~\cite{LSTMCF} are constrained by hardware limitations in actual scenarios, which may hinder their effective calculation.
\begin{figure*}[!t]
	\centering
	\includegraphics[width=5.8in]{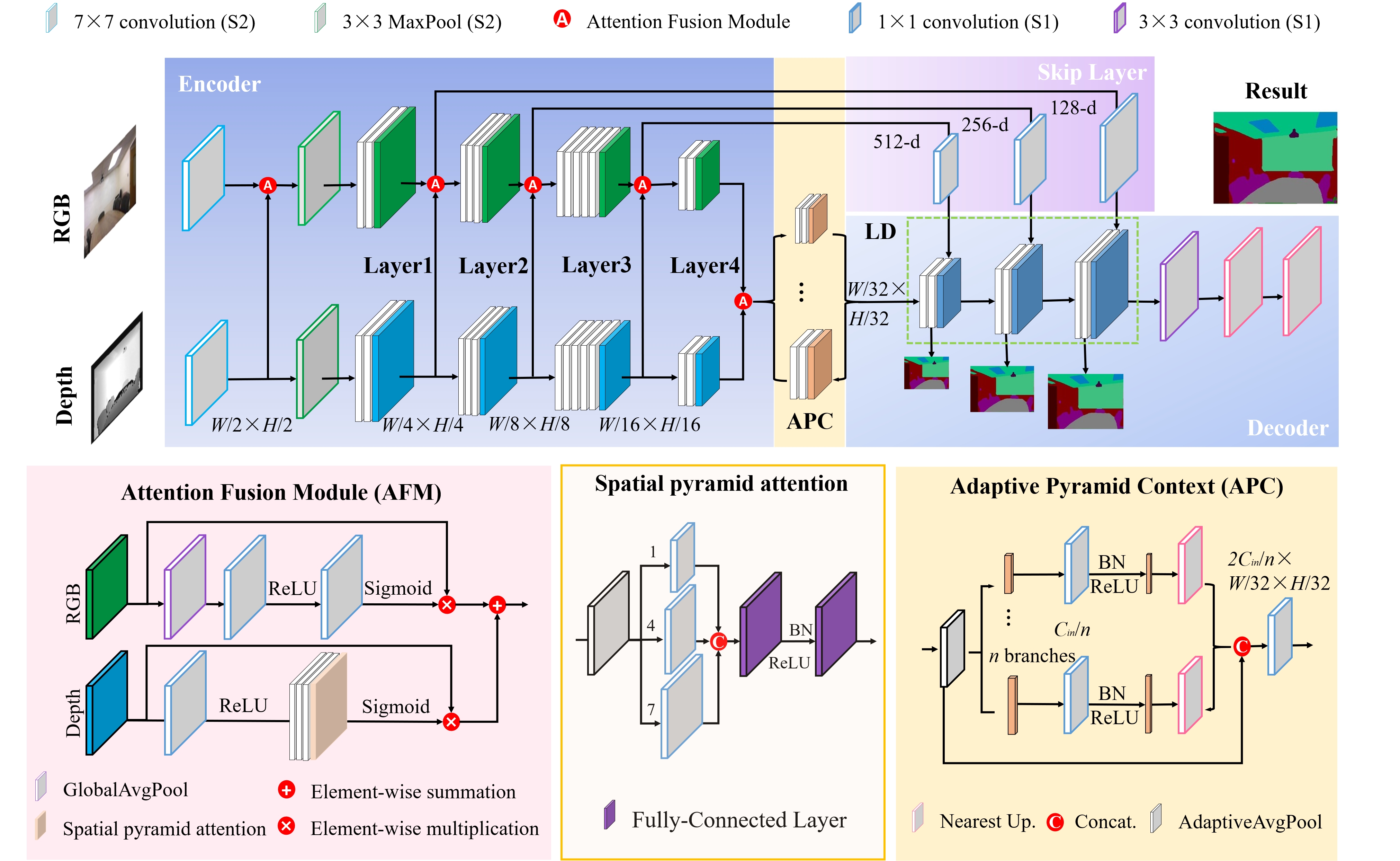}
	\caption{Overview structure of our proposed SGACNet network (top) and functional subnetworks (bottom). The SGACNet belongs to an encoder-decoder network structure. To obtain more interesting features from channel and space, the attention fusion module (AFM) is added after each encoder part. The adaptive pyramid context (APC) module utilizes global-guided local affinity to enlarge the receptive field. The decoder part receives information from skip layers and the APC module. Based on this, we output the features of each light-weighted decoder to observe and avoid the gradient disappearance problem.}
	\label{Framework}
\end{figure*}

\subsection{Efficient Semantic Segmentation}
Although depth information is significant in improving the accuracy of semantic segmentation, it increases computational complexity and thus reduces inference speed. To optimize segmentation, recent efficient methods have proposed tailored frameworks that strive to decrease parameters and calculations. Paszke \textit{et al.}~\cite{ESPNetv2} thought that replacing traditional convolution with lightweight convolution such as depth-wise separable convolution or pointwise convolutions, which can speed up the encoder while minimizing memory usage. But extensive usage of pointwise convolution in networks was considered a computational bottleneck. Zhou \textit{et al.}~\cite{ICNet} involved low, medium, and high-resolution images for cascade fusion trained with labels. Nonetheless, the intuitive speed-up strategies (\textit{e.g.} downsampling) shrunk feature maps and conducted model compression making an attribute to time reduction,  which would cause coarse prediction maps.  Oršic \textit{et al.}~\cite{SwiftNet} enlarged the receptive field by connecting the features of the encoder with upsampling and fusing the features of each resolution, whereas it demands a large number of labeled images required for fine-tuning and training.
Cao \textit{et al.}~\cite{ ShapeConv } proposed a special shape-aware convolutional layer that takes into account the shape of objects in the scene rather than simply treating them as flat 2D images. While the data preprocessing requirements are high, and the shape data needs to be preprocessed and standardized. Self-attention mechanisms~\cite{TSNet} integrated into separate encoder-decoder architectures can capture global context information, Yu \textit{et al.}~\cite{Lite-HRNet} learned the weight from each branch and different resolutions, and it applied the weight to information transmission. These networks still require substantial computational resources to train and use effectively. For this reason, one of the major directions today is making segmentation both precise and fast.

\subsection{Attention Mechanism}
Attention mechanisms come in many forms and only focus on critical features. For instance, Hu \textit{et al.}~\cite{ACNet} proposed an attention complementary module according to channel information. Semantic with geometric information was weighted and then fused to obtain higher quality. Similarly, with the spatial-channel co-attention module, Du \textit{et al.}~\cite{ Gated_Fusion } could selectively collect and fuse global bimodal features contributing to high-resolution semantic prediction. In the computing process, there are often noisy signals between different modalities. Zhu \textit{et al.}~\cite{ CMANet } suppressed noise messages from depth data by adjusting the attention mechanism. To make its edge-aware predictions more refined, Zou \textit{et al.}~\cite{ GED } employed gate-guided edge distillation for extracting information from multi-layer features.  As the attention weights represent the importance of the fusion feature, the larger the weight, the more concentrated the corresponding value. 
By adjusting the attention mechanism, it is possible to optimize the fusion process and achieve better results. However, finding the optimal distribution of weights is a complex task that requires careful analysis. Therefore, it remains a thought-provoking question in this field.
\begin{figure}[!t]
	\centering 	
	\subfigure{
		\includegraphics[width=3.2in]{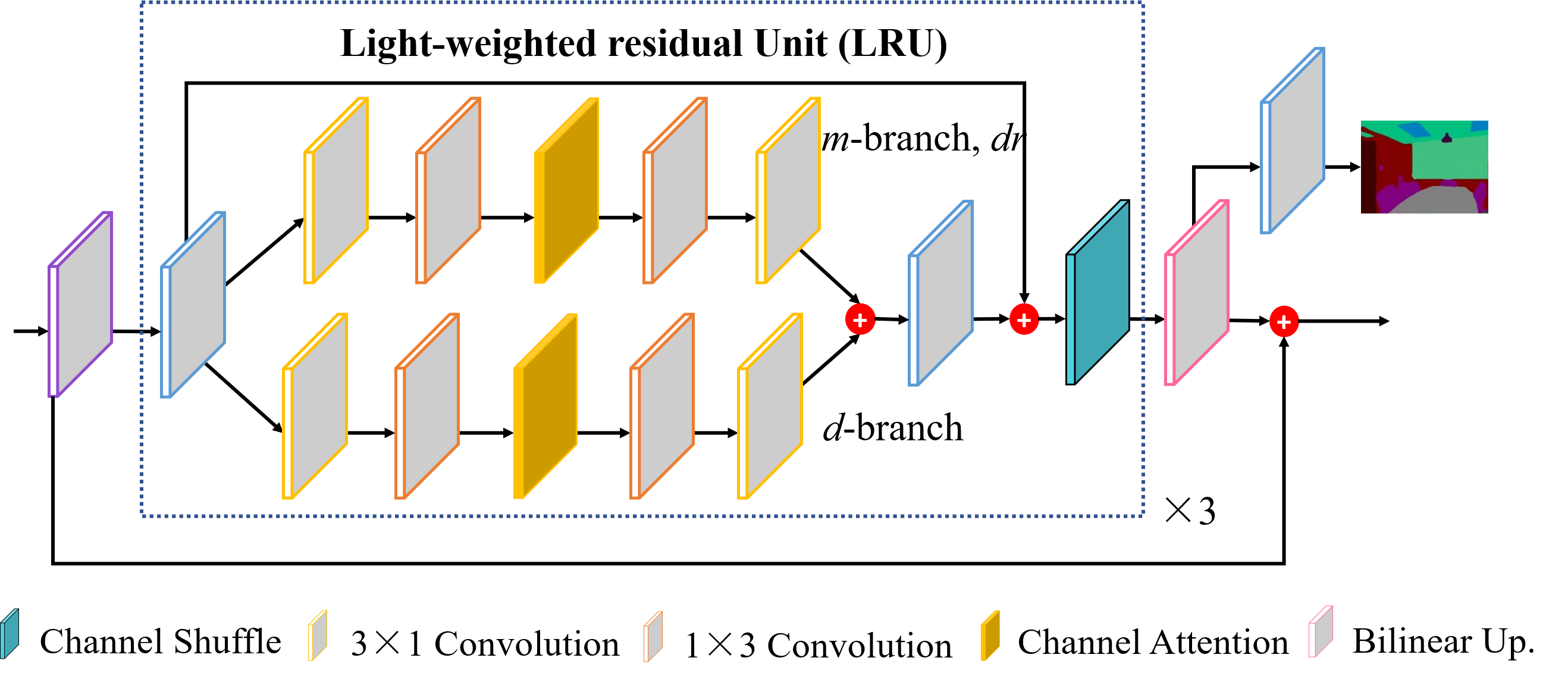}
	}
	\caption{Detailed structure of light-weighted decoder (LD). To reduce the computational cost, \textit{d}-branch takes advantage of asymmetric convolutions. It absorbs short-distance features and complements information continuity. Similarly, \textit{m}-branch is designed for extracting long-distance features and enlarging receptive fields with dilated depthwise separable convolutions, \textit{dr} means dilated rate.}
	\label{spatial}
\end{figure}

To sum up, robots need to pay attention to real-world task requirements without interference and have the ability to process information quickly. However, with limited computing resources, the existing methods are hard to implement RGB-D semantic segmentation effectively. Hence we propose our own segmentation framework by studying the previous semantic analysis and efficient segmentation method. In addition, we also discuss the role of enhanced depth information in guiding image messages and the effect of adaptive context in capturing global information.

\section{Proposed Method}\label{sec:Proposed Method}
Our network follows a classical encoder-decoder structure to reduce the degradation problem in deep CNNs. As illustrated in Fig.~\ref{Framework}, SGACNet has two separate branches in the encoder: the upper branch is used to extract RGB features, and the bottom for depth. At each layer in both branches, input features are downsampled to obtain a compressed feature map. After downsampling, depth features are selectively fused with RGB features through attention fusion modules (AFM), which help to leverage useful depth information. To make full use of global information after the encoder, we employ an adaptive pyramid context (APC) module. In the decoder, our proposed light-weighted decoder (LD) performs upsampling and utilizes the fusion features obtained from the AFM through skip connections. We also design a lightweight residual unit (LRU) to improve inference speed and performance.

\subsection{Encoder for Spatial Information Mining}

As noted previously, We are committed to designing an effective semantic segmentation architecture.  And as the encoder layer becomes deeper, the complexity of the network increases, resulting in a larger number of parameters and computations being processed. This leads to a decrease in inference speed. Therefore, we replace Non-Bottleneck-1D (NBt1D) to replace the bottleneck in ResNet~\cite{EMSANet}, achieving a balance between accuracy and inference time.
\begin{figure}[!t]
	\centering
	\subfigure[RGB]{
		\includegraphics[width=1.1in]{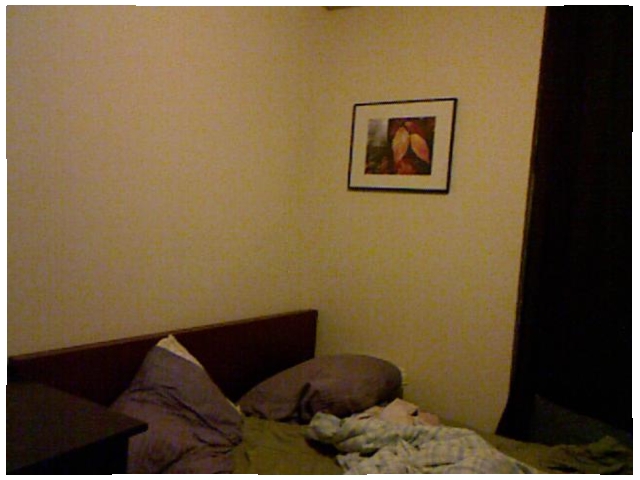}
	}\hspace{-0.8em}
	\subfigure[Layer4\_rgb\_fm]{
		\includegraphics[width=1.1in]{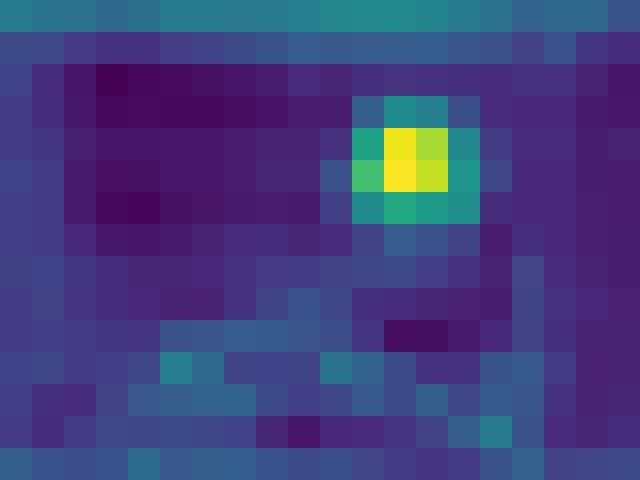}
	}\hspace{-0.8em}
	\subfigure[Layer4\_rgb\_am]{
		\includegraphics[width=1.1in]{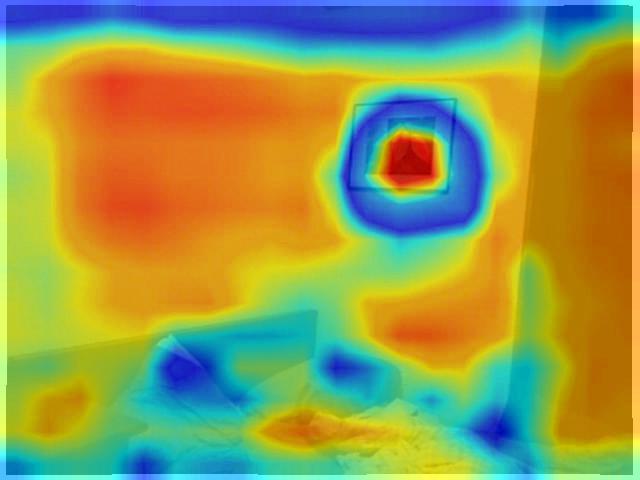}
	}\hspace{-0.8em}
	\subfigure[Depth]{
		\includegraphics[width=1.1in]{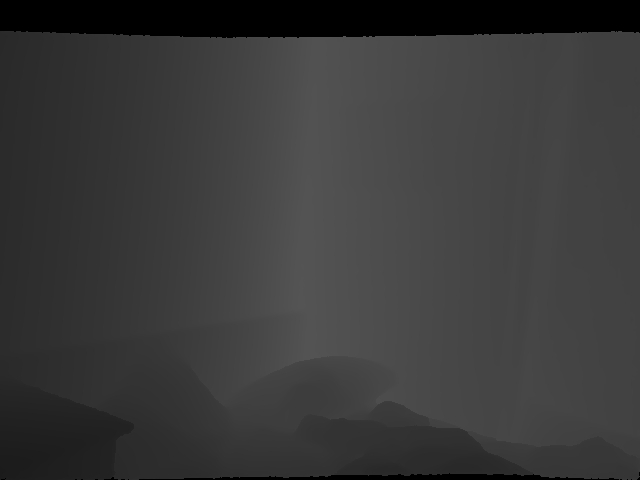}
	}\hspace{-0.8em}
	\subfigure[Layer4\_depth\_fm]{
		\includegraphics[width=1.1in]{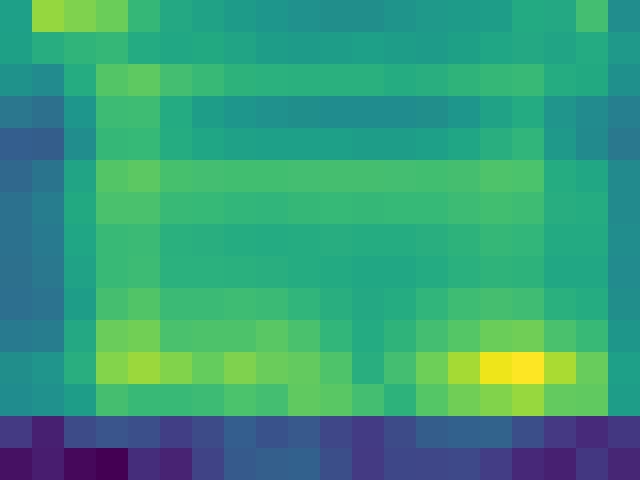}
	}\hspace{-0.8em}
	\subfigure[Layer4\_depth\_am]{
		\includegraphics[width=1.1in]{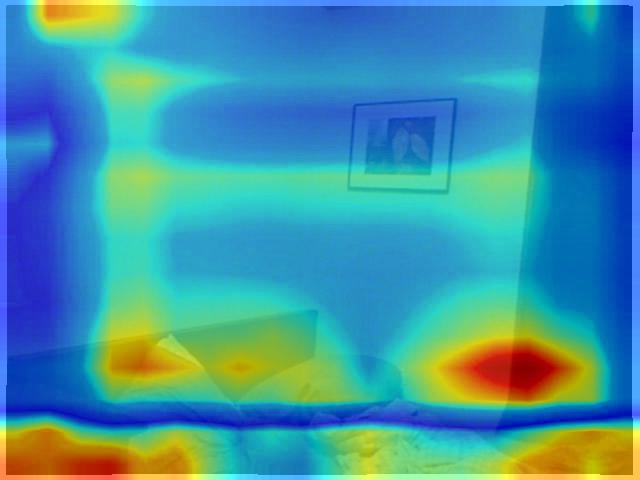}
	}\hspace{-0.8em}
	\subfigure[Result]{
		\includegraphics[width=1.1in]{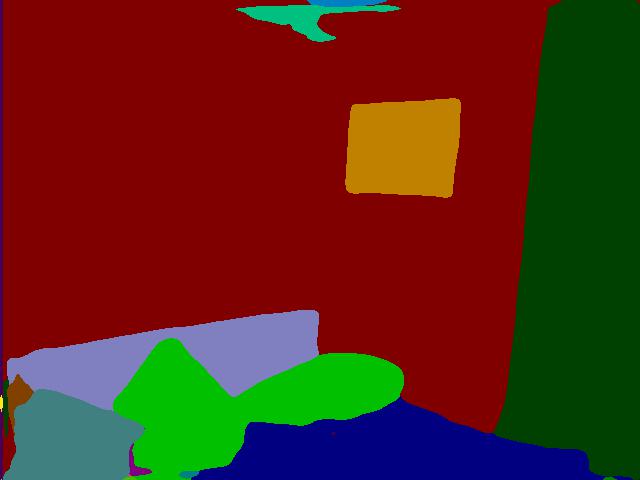}
	}\hspace{-0.8em}
	\subfigure[Layer4\_fm]{
		\includegraphics[width=1.1in]{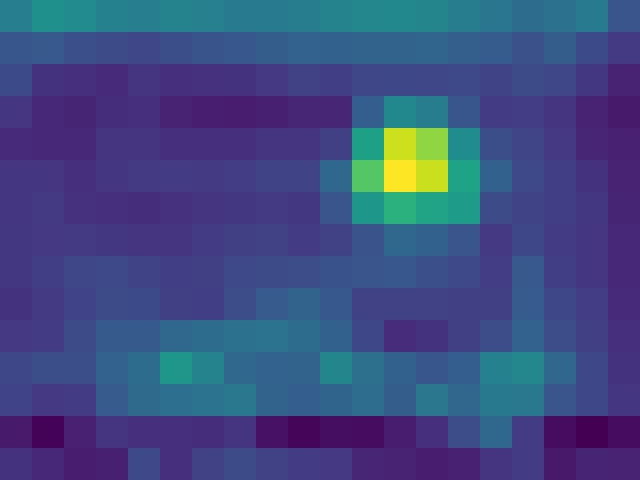}
	}\hspace{-0.8em}	
	\subfigure[Layer4\_am]{
		\includegraphics[width=1.1in]{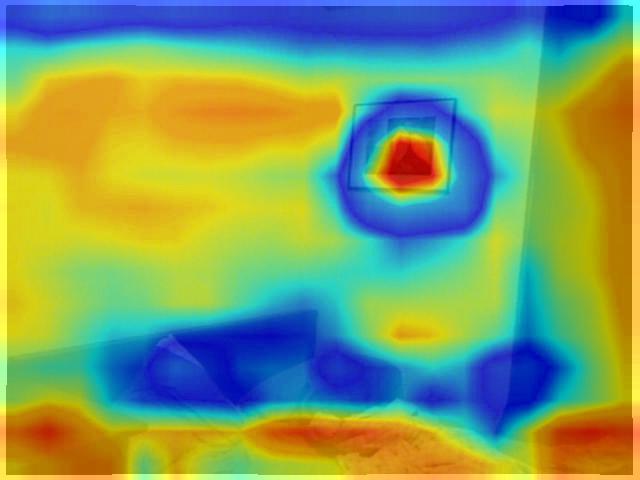}
	}
	\caption{Visual results of fusion analysis on the 4-th encoder layer. The second column is the feature map (fm), and the last column is the attention map (am). Layer4\_fm/am is the fusion result of Layer4\_rgb\_fm/am and Layer4\_depth\_fm/am. Attention weights are from feature maps.}
	\label{fig:attention}
\end{figure}

From Fig.~\ref{fig:attention}, we observe that RGB and depth maps capture different feature expressions. RGB focuses on contour information, and depth can express location information more clearly without being influenced by light and shadow conditions. To address the issue of uneven information distribution among channels, we propose a fusion module that employs two attention mechanisms to filter out insignificant features. The Fig.~\ref{fig:attention} shows that our AFM does work well in balancing multimodal features.
We adopt $\mathop X$ = $\left[ {{X_1},{X_2}, \ldots ,{X_C}} \right]$ $\in$ ${\mathbb{R}^{C \times H \times W}}$ as input feature map, $
C, H, W
$ are the channel numbers, height, and width of the input, respectively.
To apply the channel attention mechanism, we first reduce the spatial dimensions of each feature map by global-average pooling to obtain the \textit{l}-th vector $Q\left( {{x_l}} \right)$ $\in$ ${\mathbb{R}^{C \times 1 \times 1}}$, following as
 \begin{equation}
 		\mathop Q\left( {{x_l}} \right) = \frac{1}{{H \times W}}\sum\limits_i^H {\sum\limits_j^W {{{x_l}}\left( {i,j} \right)} }.
 \end{equation}

To improve recognition accuracy, spatial pyramid attention transforms features of different scales by adaptive pyramid pooling shown in Fig.~\ref{Framework}. The \textit{C$\left( { \cdot  } \right)$} and $P\left( { \cdot  } \right)$ are denoted as a concatenation layer and a pooling operator, respectively. The $R_{i^{\prime}}(\cdot)$ means vector resizing operation by standard convolutional layers of kernel size $i^{\prime}$. The attention pyramid can be presented as
\begin{equation}
 S\left( {{x_l}} \right) =C\left( {{R_7}\left( {P\left( {{x_l}} \right)} \right),{R_4}\left( {P\left( {{x_l}} \right)} \right),{R_1}\left( {P\left( {{x_l}} \right)} \right)} \right).
\end{equation}
 
Then, the above operators need a fully connected layer ${F_f}\left(  \cdot  \right)$ to learn and connect their previous local classification information. We follow activate layers as ReLU function $\delta$ and sigmoid $\sigma$. To obtain final attention results, we perform an element-wise multiplication ${M_k}\left( { \cdot} \right) $, $k=c$ represents the channel branch, and $k=s$ represents the spatial branch. A fusion feature map is generated as
\begin{equation}
sum\left({{x_l}} \right) = {M_c}\left( {{x_l}} \right) \oplus {M_s}\left( {{x_l}} \right),
\end{equation}
where 
\begin{equation}
	{M_k}\left( {{x_l}} \right) = {x_l} \sigma \left( {{F_f}\left( {{F_f}\left( {{U_k}} \right)} \right)} \right),
\end{equation}
\begin{equation}
U_k= \begin{cases}Q\left(x_l\right), & k=c \\ S\left(x_l\right), & k=s\end{cases}
\end{equation}

Actually, there are various spatial attention modules proposed recently. More selective propositions are discussed and parts of them are provided in Section~\ref{sec:Ablation}.

\subsection{Adaptive Pyramid Context  Module}
During the encoding and decoding process, certain operations, such as upsampling and pooling may result in the loss of important image information. Therefore, the context module is essential for enhancing image features. As the input features have different resolutions at different scales, a feature pyramid can obtain information about the desired target according to the corresponding scale, thereby improving the performance of the entire feature map.

\begin{figure}[!t]
	\centering
	\subfigure[Inputs]{
		\includegraphics[width=1.04in]{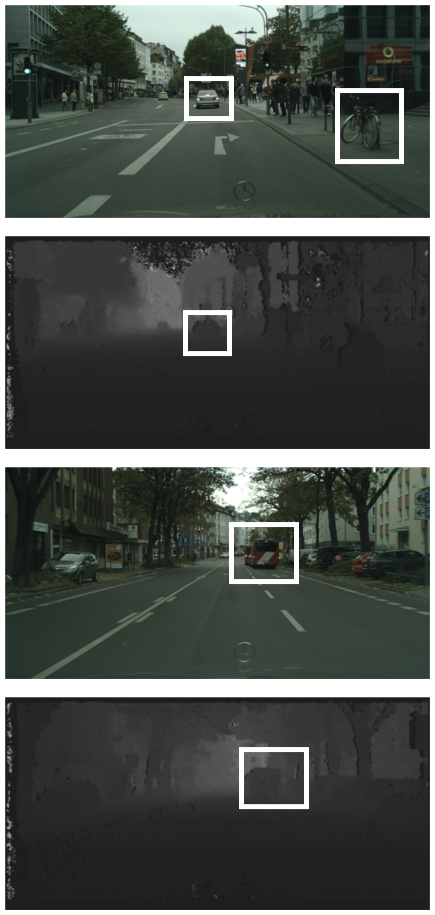}
	}\hspace{-0.5em}
	\subfigure[w/o APC]{
		\includegraphics[width=1.04in]{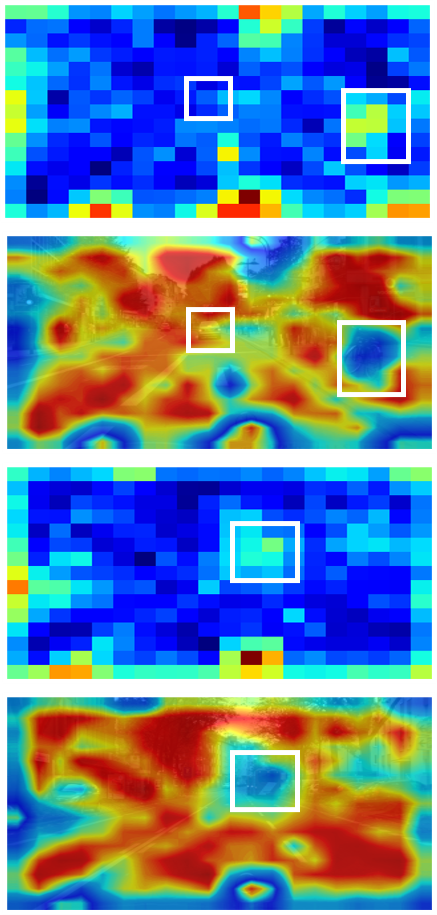}
	}\hspace{-0.5em}
	\subfigure[w/ APC]{
		\includegraphics[width=1.04in]{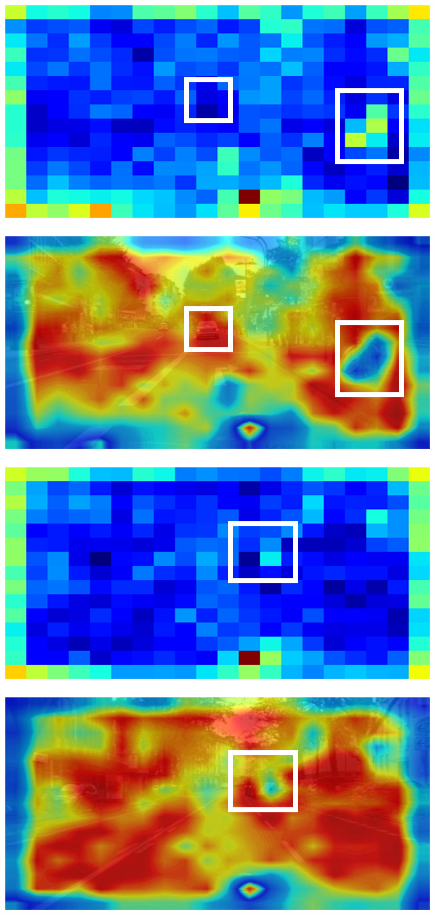}
	}
	\caption{Visual comparison of context analysis on Cityscapes dataset. Attention maps and feature maps are on the same line with RGB and depth input. (b) and (c) are visualizations without and with our proposed adaptive pyramid context modules. Attention weights are from feature maps. Warmer colors represent the areas that receive more attention, while cooler colors represent the areas with less attention.}
	\label{context}
\end{figure}

Combining the above characteristics, our network incorporates a global-guided local affinity adaptive semantic module, similar to \cite{APCNet}, as shown in Fig.~\ref{Framework} (bottom right). This module integrates information of different scales, with the number of branches changing based on variable conditions. In Fig.~\ref{context}, we visualize the comparison of context analysis,  showing that our proposed context layer provides a more accurate recognition range compared to without it. In other words, contextual information with a larger receptive field helps to understand shared features in near local regions of different categories, thus improving the segmentation performance. To further reduce the computational complexity, we use the nearest upsampling operation, which has a simple structure.  

\subsection{Light-weighted Decoder}
The calculation is a key factor that hinders model deployment. Depthwise separable convolution has smaller computations, but it needs to save more intermediate variables, which leads to longer read-and-write time and slower training speed. Atrous convolution is often used in real-time tasks to provide a larger receptive field with the same amount of calculation. However, it also leads to problems such as loss of information continuity and irrelevance of long-distance information, which can be fatal for pixel-level semantic segmentation.

Therefore, we propose the LD to solve the above problems. After repeated trials, we find that the segmentation effect of the three light-weighted residual units is better. The entire structure employs a classic cross-channel model, in which the dual-branch design adopts asymmetric convolution ${A_j^{\prime}}\left(  \cdot  \right)$ (1$\times$3, 3$\times$1), $j^{\prime}$ is the number of uses. This structure can approximate the existing convolution, ensure the same amount of calculation, compress the model, and accelerate it. 
As shown in Fig.~\ref{spatial}, to supplement information continuity and obtain close-range feature information, the \textit{d}-branch ${z_{d,k}}$ adopts depth-wise separable convolution, the value $k$ indicates how many times the feature map has passed through unpaired convolutions. The intermediate branch ${z_{m,k}}$ adopts dilated convolutions based on the \textit{d}-branch to reduce computational cost and obtain deeper network features. 
\begin{equation}
	{z_{\_,2}}\left( {{y_l}} \right) = {A_2}\left( {{V_{ac}}\left( {{A_1}\left( {{W_1}\left( {{y_l}} \right)} \right)} \right)} \right),
\end{equation}
where ${y_l}$ is the $l$-th decoder feature map assumed as $\mathop Y$ = $\left[ {{y_1},{y_2}, \ldots ,{y_C}} \right]$ $\in$ ${\mathbb{R}^{C \times H \times W}}$. The LD module integrates features from long and short distances shown as
\begin{equation}
Z\left( {{y_l}} \right) = f\left( {{W_1}\left( {{z_{d,2}}\left( {{y_l}} \right) \oplus {z_{m,2}}\left( {{y_l}} \right)} \right) \oplus {W_1}\left( {{y_l}} \right)} \right),
\end{equation}
where ${W_{1}}$ means the weight of a 1$\times$1 convolutional layer. $Z\left(  \cdot  \right)$ represents the output of the down or mid branch. After making summation among these branches, a channel shuffle $f\left(  \cdot  \right)$  has realized features communication among them.

We also use the channel attention module (CAM) ${V_{ac}}\left( . \right)$ to enhance semantic expression
\begin{equation}
	\mathop V\nolimits_{ac}  = \left[ {\sigma \left( {\mathop {{y_1}}\limits^ \wedge  } \right)\mathop y\nolimits_1 ,\sigma \left( {{{\mathop y\limits^ \wedge  }_2}} \right)\mathop y\nolimits_2 ,\; \ldots ,\sigma \left( {{{\mathop y\limits^ \wedge  }_n}} \right)\mathop y\nolimits_n } \right].
\end{equation}
The input messages $\mathop y\limits^ \wedge   = {W_1}\left( {Q\left( {{y_l}} \right)} \right)$ are firstly operated by a global average pooling and activated by sigmoid $\sigma$ later.

The feature maps of the last attention fusion layer depict images being continuously downsampled, resulting in fairly blurred pixels. To effectively integrate features from relevant image regions and corresponding pixel semantic labels, we introduce the adaptive pyramid context module (APC). For a visual comparison of the results in Table 4, we encircle the areas with obvious differences using white boxes, such as cars and buses on the road, and bicycles. It is evident that our method with the APC can focus on capturing more detailed multi-scale context information, leading to improved segmentation performance. 

Besides, we take the training output as the input of the cross-entropy (CE) function for loss calculation
\begin{equation}
{L_{p,q}} =  - \sum\limits_{i = 1}^n {{p_i}} \log \left( {{q_i}} \right),
\end{equation}
where $n$ represents the number of categories. The ${p_i}$ is the real, and ${q_i}$ is the prediction. As a matter of fact, CE loss is classic for classification and prediction. Because when deriving the gradient ${h}$ of loss function ${\frac{{\partial L}}{{\partial {h_i}}} = {q_i}\left( {1 - {q_i}} \right)}$, the output value of back-propagation weight gets closer to 0 or 1, the gradient will disappear in some cases, like mean square error. However, this does not exist in the cross-entropy function, so training is more likely to continue.

\section{Experiments}\label{sec:Experiments}
In this section, our experimental setup is first introduced including three benchmark datasets, five metric evaluation criteria, and other details. We then perform ablation experiments on the NYUv2 dataset. To verify the superiority of the proposed method, we compare it with the state-of-the-art (SOTA) methods on three datasets.
\begin{table*}[!t]
	\centering
	\renewcommand{\arraystretch}{1.2}
	\caption{Ablation study of the proposed method on NYUv2 dataset. Attention Fusion Module is a double-branch structure, SE:squeeze-and-excitation attention. SPGE:spatial-group-enhance attention, SPA147:spatial pyramid attention, and the number in SPA147 represent the bin sizes of average pooling. PPM: pyramid pooling model. APC: adaptive pyramid context module. NDM: normal decoder module. LD: light-weighted decoder.}
	\label{table1}
	\footnotesize
	\begin{tabular}{ccccccccccc}
	\toprule 
	\multicolumn{1}{c}{\multirow{2}{*}{\textbf{Upsample}}}&\multirow{2}{*}{\textbf{Context}}	& 	\multirow{2}{*}{\textbf{Decoder}} &
		\multicolumn{5}{c}{\textbf{Attention Fusion Module}}  	&
		\multicolumn{3}{c}{\textbf{ResNet18}} \\
		\cmidrule(r){4-8} \cmidrule(r){9-11}
\multicolumn{1}{c}{} & & &\multicolumn{1}{c}{\textbf{\textrm{2$\times$SE}}}            			&
		\multicolumn{1}{c}{\textbf{\textrm{2$\times$SPGE}}}            			&
		\multicolumn{1}{c}{\textbf{\textrm{2$\times$SPA147}}}            		&
		\multicolumn{1}{c}{\textbf{\textrm{SE$+$SPGE}}}            		&
		\multicolumn{1}{c}{\textbf{\textrm{SE$+$SPA147}}}  &\multicolumn{1}{c}{\textbf{$\textrm{mIoU}\uparrow$}}&\textbf{$\textrm{FPS}\uparrow$} &\textbf{$\textrm{Params}\downarrow$}      \\
    \midrule
		L3$\times$3& PPM &NDM &\multicolumn{1}{c}{\multirow{1}{*}{$\checkmark$}}&  &  &   &  &47.9&15.4 &33.5 \\
		L3$\times$3& PPM &NDM& &\multicolumn{1}{c}{\multirow{1}{*}{$\checkmark$}} &    &    &    &47.8 &16.0 &33.4\\
		L3$\times$3& PPM &NDM& & &\multicolumn{1}{c}{\multirow{1}{*}{$\checkmark$}}&    & &47.9 &16.6&33.5 \\
	    L3$\times$3&PPM&NDM& & & &\multicolumn{1}{c}{\multirow{1}{*}{$\checkmark$}}&    &48.3&17.0&33.4 \\
		L3$\times$3&PPM&NDM&&&&&\multicolumn{1}{c}{\multirow{1}{*}{$\checkmark$}}&48.6&17.4&33.5\\
    \midrule
		L3$\times$3&APC&NDM&&&&\multicolumn{1}{c}{\multirow{1}{*}{$\checkmark$}}&    &48.4 &16.2&33.4 \\	
		L3$\times$3 &APC&NDM& & && &\multicolumn{1}{c}{\multirow{1}{*}{$\checkmark$}}&48.7&16.7&33.5\\
    \midrule 
		L3$\times$3&APC&LD&&&&\multicolumn{1}{c}{\multirow{1}{*}{$\checkmark$}}&&47.5&15.4&22.1 \\
	  L3$\times$3&APC&LD&&&&&\multicolumn{1}{c}{\multirow{1}{*}{$\checkmark$}}&47.9&15.8&22.2\\
		Bilinear &APC&LD&&&& &\multicolumn{1}{c}{\multirow{1}{*}{$\checkmark$}}&48.2&15.9&22.2 \\
	\bottomrule
		\label{tab:AFM}
	\end{tabular}
\end{table*}

\subsection{Experimental setup}
To evaluate the performance of our proposed approach, we conduct the experiments on three RGB-D datasets, namely NYUv2 \cite{NYUv2}  and SUN RGB-D \cite{SUNRGB-D} for indoor scenes and Cityscapes \cite{Cityscapes} dataset for outdoor scenes. 
\begin{figure*}[!t]
	\centering
	\subfigure[RGB]{
		\includegraphics[width=0.95in]{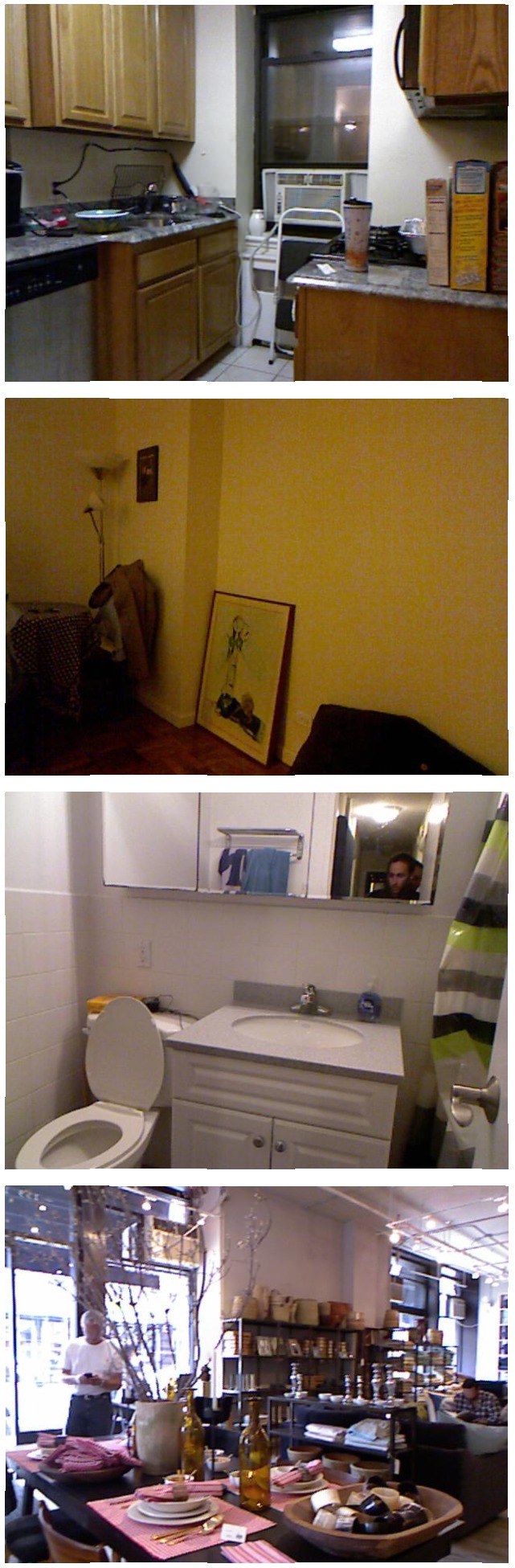}
	}\hspace{-0.60em}
	\subfigure[Depth]{
		\includegraphics[width=0.95in]{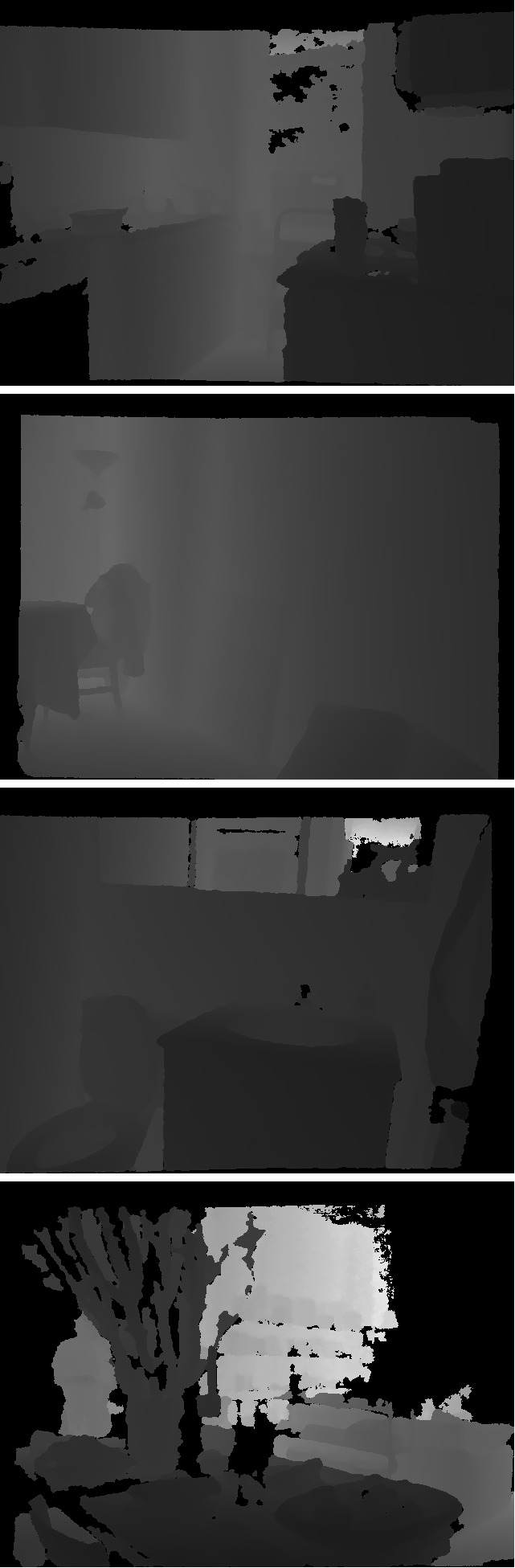}
	}\hspace{-0.55em}
	\subfigure[2$\times$SE]{
		\includegraphics[width=0.95in]{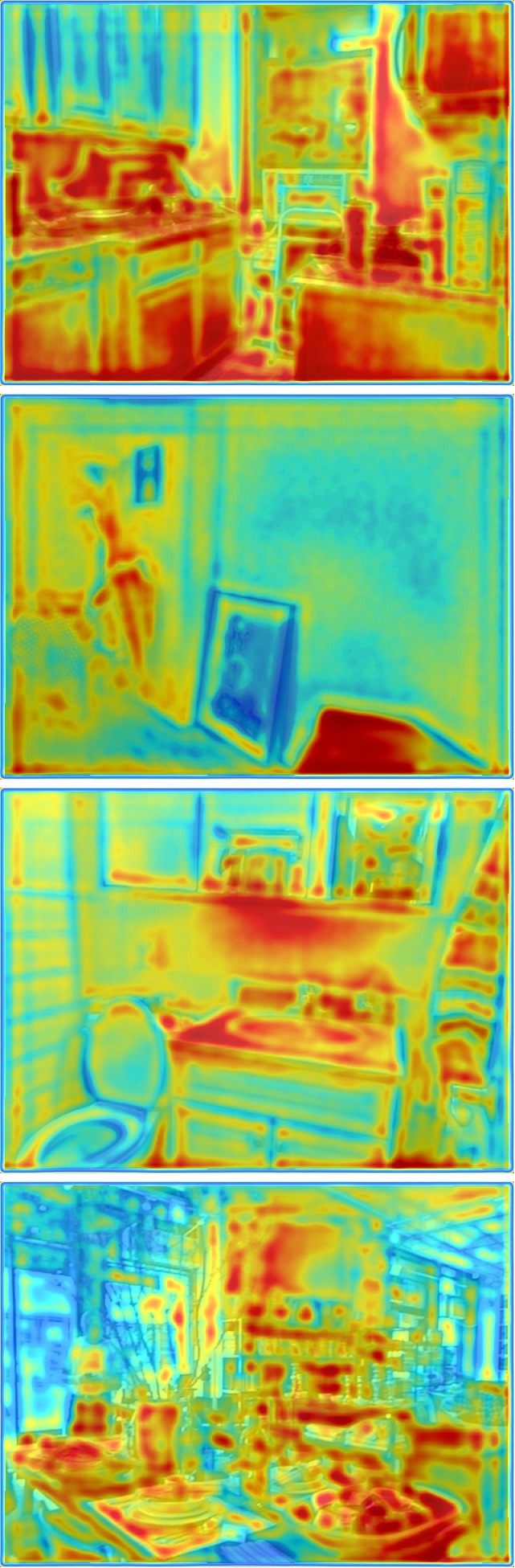}
	}\hspace{-0.55em}
	\subfigure[2$\times$SPGE]{
		\includegraphics[width=0.95in]{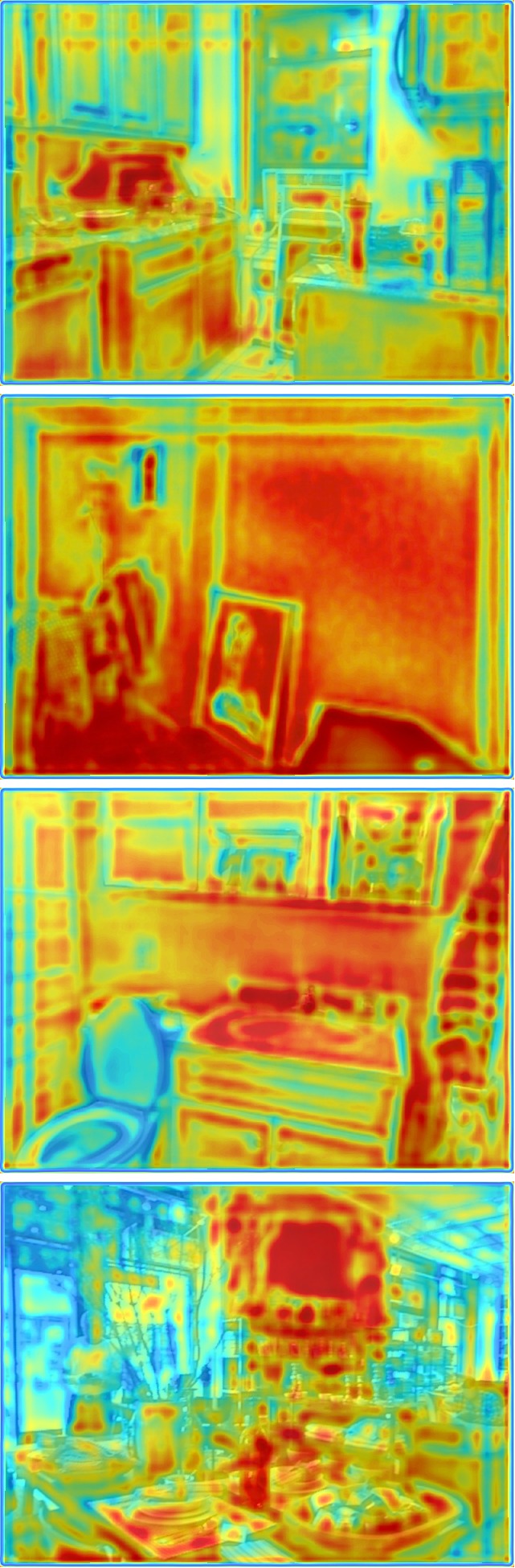}
	}\hspace{-0.55em}
	\subfigure[2$\times$SPA147]{
		\includegraphics[width=0.95in]{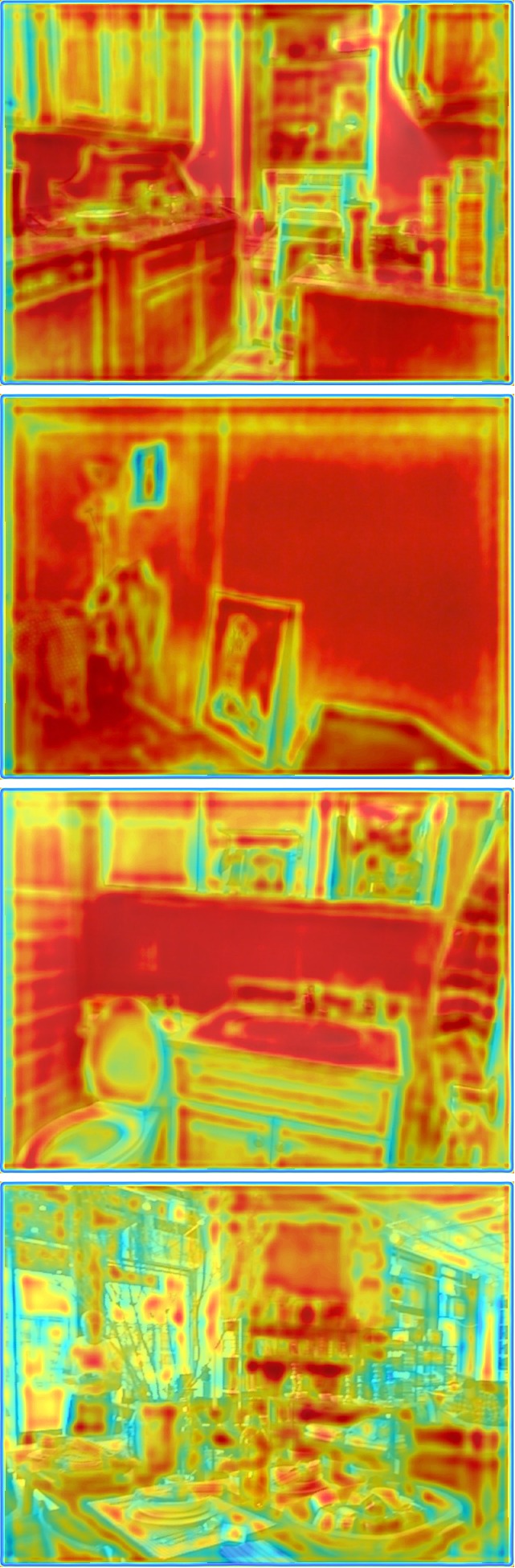}
	}\hspace{-0.55em}
	\subfigure[SE$+$SPGE]{
		\includegraphics[width=0.95in]{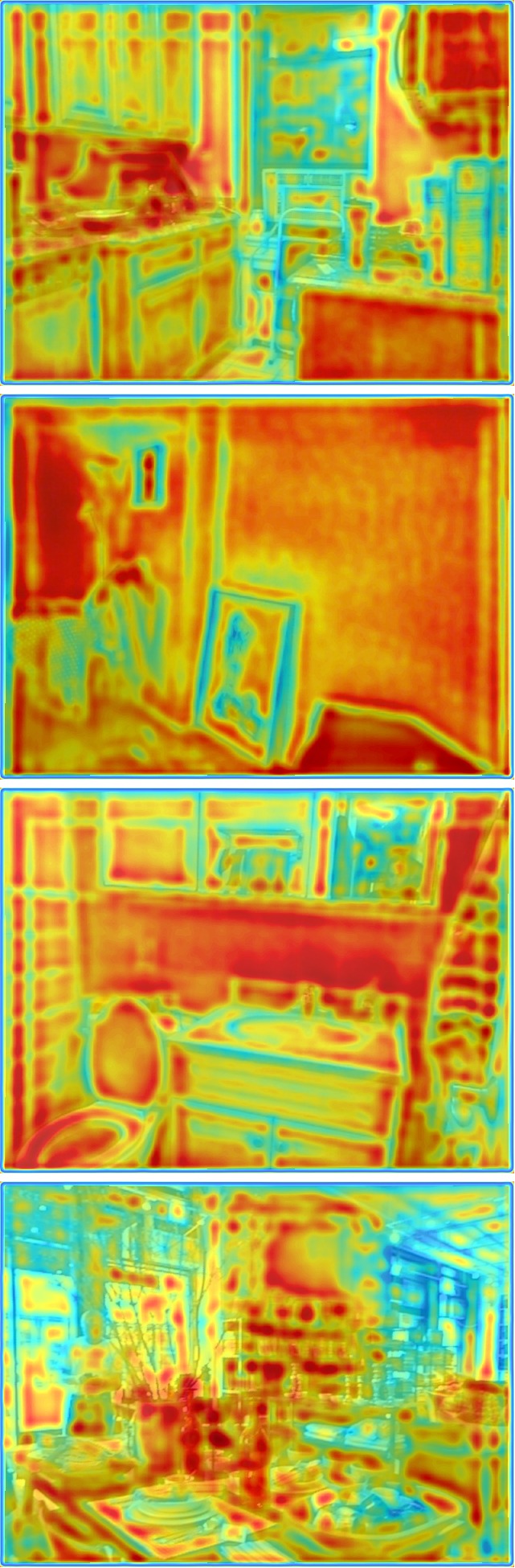}
	}\hspace{-0.55em}
	\subfigure[SE$+$SPA147]{
		\includegraphics[width=0.95in]{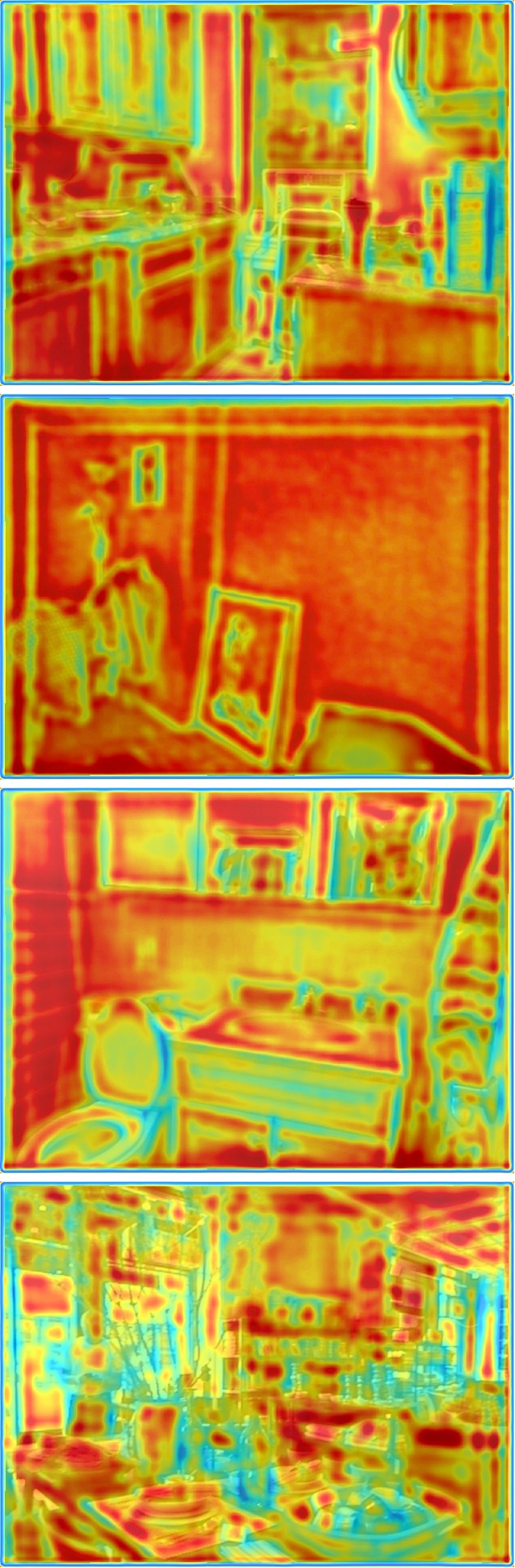}
	}
	\caption{The visual results of attention fusion module on NYUv2 dataset. These results are calculated for the last decoder convolution. The smaller the weight distribution value in space, the closer to blue, like the contour of an object, and the larger to red.}
	\label{AFM}
\end{figure*}

\subsubsection{Indoor datasets}
NYUv2 consists of 1449 RGB-D images, of which the standard training and testing sets are split into 795 and 654 images, respectively. And there are 40 common class labels. The SUN RGB-D with 37 classes contains 10335 RGB-D images. There are 5285 images from the official training set for training our network and the official testing set with 5050 images for evaluation. In particular, we resized the inputs to a resolution of 640×480 pixels for the above two datasets. In addition, to avoid over-fitting, we augment the images with strategies like random scaling, horizontal flipping, and random cropping. 

\subsubsection{Outdoor datasets}
Cityscapes dataset contains 5000 images with a high resolution of 1024$\times$2048 pixels with fine-grained annotation for 19 classes. We use 2975 images for training, 500 for validation, and 1525 for testing. Because We also consider efficient semantic segmentation, the network input resolution is set to 512$\times$1024 pixels.

\subsubsection{Metrics}
For the evaluation of efficient semantic segmentation, there are five common metrics for evaluation, including mean intersection over union (mIoU), pixel accuracy (PixAcc.), mean accuracy (mAcc.) for accuracy segmentation, frame per second (FPS) for inference speed, and space complexity (Params) of a different model.

\subsubsection{Implementation details}
Based on PyTorch1.3, CUDA V10.1, and Python 3.7, We train our method for 500 epochs with 8 batch sizes. The optimizer used for NYUv2 and SUN RGB-D datasets is Stochastic Gradient Descent (SGD) with a momentum value of 0.9. For the Cityscapes dataset, the Adam optimizer is used, with learning rates of {0.00125, 0.0025, 0.005, 0.01, 0.02, 0.04} and {0.0001, 0.0004}, respectively, and a weight decay of 0.0001.  We adapt the learning rate with one cycle learning rate scheduler~\cite{EMSANet}. 

\begin{table*}[t]
	\renewcommand\arraystretch{1.08}
	\begin{center}
		\setlength{\belowcaptionskip}{1cm}
		\caption{Each category on the NYUv2 is compared with the state-of-the-arts. Percentage (\%) of IoU is displayed for evaluation, with the best performance denoted in \textbf{bold}.}
		\label{tab:imbalance1}
		\resizebox{\textwidth}{!}{
			\begin{tabular}{lccccccccccccccccccccc}
				\toprule
				Method & \rotatebox{90}{wall} & \rotatebox{90}{floor} & \rotatebox{90}{cabinet} & \rotatebox{90}{bed} & \rotatebox{90}{chair} & \rotatebox{90}{sofa} & \rotatebox{90}{table} & \rotatebox{90}{door} & \rotatebox{90}{window} & \rotatebox{90}{bookshelf} & \rotatebox{90}{picture} & \rotatebox{90}{counter} & \rotatebox{90}{blinds} & \rotatebox{90}{desk} & \rotatebox{90}{shelves} & \rotatebox{90}{curtain} & \rotatebox{90}{dresser} & \rotatebox{90}{pillow} & \rotatebox{90}{mirror} & \rotatebox{90}{floormat}\\
				\midrule
				DCNV2~\cite{DCNV2}         &76.8 &82.7 &56.3 &64.3 &57.0 &59.9 &39.9 &36.3 &44.5 &45.6 &60.3 &56.0 &57.1 &21.1 &16.6 &54.1 &47.3 &41.9 &36.5 &32.8 \\
				VCD~\cite{VCD}        &78.2 &83.7 &57.4 &66.1 &57.2 &60.9 &40.1 &39.5 &45.1 &46.8 &59.4 &58.1 &56.6 &21.9  &16.0  &55.2 &47.0 &42.7 &36.2 &34.3 \\
				ESANet~\cite{ESANet}              &79.9 &87.7 &62.4 &71.2 &63.6 &64.2 &43.6 &39.5 &47.6 &42.7 &61.8 &67.4 &59.7 &23.4 &17.4 &58.3 &44.9 &49.9 &50.2 &37.2 \\
                Gated Fusion~\cite{Gated_Fusion}        &78.5 &87.1 &56.6 &70.1 &65.2 &63.9 &46.9 &35.9 &47.1 &48.9 &54.3 &66.3 &51.7 &20.6 &13.7 &49.8 &43.2 &50.4 &48.5&32.2 \\
				SGACNet (Ours)                       &\textbf{80.3} &\textbf{87.7} &60.9 &\textbf{71.5} &64.6 &63.6 &45.8 &\textbf{39.9} &\textbf{49.0} &44.9 &\textbf{61.9} &67.1 &\textbf{60.1} &\textbf{24.9} &\textbf{18.8} &\textbf{60.6} &\textbf{51.2} &47.9 &49.1 &\textbf{39.4} \\
				\bottomrule
				\toprule
				Method & \rotatebox{90}{clothes} & \rotatebox{90}{ceiling} & \rotatebox{90}{books} & \rotatebox{90}{fridge} & \rotatebox{90}{tv} & \rotatebox{90}{paper} & \rotatebox{90}{towel} & \rotatebox{90}{shower} & \rotatebox{90}{box} & \rotatebox{90}{board} & \rotatebox{90}{person} & \rotatebox{90}{nightstand} & \rotatebox{90}{toilet} & \rotatebox{90}{sink} & \rotatebox{90}{lamp} & \rotatebox{90}{bathtub} & \rotatebox{90}{bag} & \rotatebox{90}{ot. struct.} & \rotatebox{90}{ot. furn.} & \rotatebox{90}{ot. props.}\\
				\midrule
				DCNV2~\cite{DCNV2}
                &21.0 &64.2 &29.1 &54.3 &60.2 &27.8 &38.1 &39.7 &13.0  &65.5 &76.1 &43.6 &66.2 &48.8 &40.0 &40.6 &10.1  &26.9 &16.2  &36.0 \\
				VCD~\cite{VCD}         
                &22.2 &67.0 &30.0 &50.9 &57.0 &30.7 &36.7 &40.6 &15.6 &72.6 &77.5 &41.2 &69.1 &51.8 &43.0 &39.4 &9.5  &27.7 &18.3  &37.0 \\
				ESANet~\cite{ESANet}               &26.2 &75.0 &32.8 &54.6 &58.0 &32.0 &38.7 &39.1 &8.4 &60.0 &77.5 &46.6 &75.3 &65.6 &52.0 &49.3 &8.7 &31.4 &20.5 &39.0 \\
                Gated Fusion~\cite{Gated_Fusion}        &24.7 &62.0 &34.2 &45.3 &53.4 &37.7 &42.6 &23.9 &11.2 &58.8 &53.2 &54.1 &80.4 &59.2 &45.5 &52.6 &15.9 &12.7 &26.4&29.3 \\
				SGACNet (Ours)                        &21.5 &\textbf{76.4} &32.6 &52.0 &50.3 &33.0 &40.6 &\textbf{43.2} &11.5 &57.1 &76.7 &49.3 &76.5&\textbf{65.7} &51.4 &\textbf{54.9} &9.8  &\textbf{31.9} &21.1 &\textbf{39.2} \\
				\bottomrule
		\end{tabular}}
	\end{center}
\end{table*}
\begin{table*}[t]
	\renewcommand\arraystretch{1.08}
	\begin{center}
		\caption{Each category on the SUN RGB-D is compared with the state-of-the-art. Percentage (\%) of IoU is displayed for evaluation, with best performance denoted in \textbf{bold}.}
		\label{tab:imbalance2}
		\resizebox{\textwidth}{!}{
			\begin{tabular}{lccccccccccccccccccccc}
				\toprule
				Method & \rotatebox{90}{wall} & \rotatebox{90}{floor} & \rotatebox{90}{cabinet} & \rotatebox{90}{bed} & \rotatebox{90}{chair} & \rotatebox{90}{sofa} & \rotatebox{90}{table} & \rotatebox{90}{door} & \rotatebox{90}{window} & \rotatebox{90}{bookshelf} & \rotatebox{90}{picture} & \rotatebox{90}{counter} & \rotatebox{90}{blinds} & \rotatebox{90}{desk} & \rotatebox{90}{shelves} & \rotatebox{90}{curtain} & \rotatebox{90}{dresser} & \rotatebox{90}{pillow} & \rotatebox{90}{mirror} \\
				\midrule
				Song \emph{et al.}~\cite{SUNRGB-D}  &36.4 &45.8 &15.4 &23.3 &19.9 &11.6 &19.3 &6.0 &7.9 &12.8 &3.6 &5.2 &2.2 &7.0 &1.7 &4.4 &5.4 &3.1 &5.6\\
                LSTMCF~\cite{LSTMCF}    &74.9 &82.3 &47.3 &62.1 &67.7 &55.5 &57.8 &45.6 &52.8 &43.1 &56.7 &39.4 &48.6 &37.3 &9.6 &63.4 &35.0 &45.8 &44.5 \\
				ESANet~\cite{ESANet}      &80.1 &90.4 &50.5 &74.4 &73.6 &57.0 &54.9 &47.5 &52.4 &35.8&55.8 &37.7 &30.2 &25.1 &19.9 &60.5 &50.5 &45.2 &51.8\\
                3DGNN~\cite{3DGNN}    &80.1 &89.9 &46.8 &73.6 &72.3 &57.6 &54.6 &47.8 &50.5 &38.3 &53.4 &37.8 &37.5  &18.1 &12.5 &60.4 &45.6 &41.7 &50.3 \\
				SGACNet (Ours)                       &\textbf{80.7} &\textbf{90.6} &\textbf{50.8} &\textbf{74.6} &\textbf{73.8} &\textbf{58.1} &55.1 &\textbf{48.0} &\textbf{53.3} &34.0 &55.2 &36.8 &30.8 &17.2 &11.6 &61.6 &\textbf{52.0} &43.6 &\textbf{51.9} \\
				\bottomrule
				\toprule
				Method & \rotatebox{90}{floormat} & \rotatebox{90}{clothes} & \rotatebox{90}{ceiling} & \rotatebox{90}{books} & \rotatebox{90}{fridge} & \rotatebox{90}{tv} & \rotatebox{90}{paper} & \rotatebox{90}{towel} & \rotatebox{90}{shower} & \rotatebox{90}{box} & \rotatebox{90}{board} & \rotatebox{90}{person} & \rotatebox{90}{nightstand} & \rotatebox{90}{toilet} & \rotatebox{90}{sink} & \rotatebox{90}{lamp} & \rotatebox{90}{bathtub} & \rotatebox{90}{bag} &\rotatebox{90}{mIoU}\\
				\midrule
				Song \emph{et al.}~\cite{SUNRGB-D}&0.0 &1.4 &35.8 &6.1 &9.5 &0.7 &1.4 &0.2 &0.0 &0.6 &7.6 &0.7 &1.7 &12.0 &15.2 &0.9 &1.1 &0.6 &9.0\\
                LSTMCF~\cite{LSTMCF}  &0.0 &28.4 &68.0 &47.9 &61.5 &52.1 &36.4 &36.7 &0.0 &38.1 &48.1 &72.6 &36.4 &68.8 &67.8 &58.0 &65.6 &23.5 &47.6\\
				ESANet~\cite{ESANet}      &0.0 &26.9 &74.4 &36.4 &55.2 &51.4 &27.2 &36.0 &22.6 &30.3 &60.4 &43.5 &24.1 &80.8 &66.0 &46.6 &69.5 &17.2&47.5 \\
				3DGNN~\cite{3DGNN} &2.5 &28.1 &71.8 &32.8 &54.4 &55.7 &23.9 &33.7 &2.7 &28.0 &54.0 &55.0 &17.2 &77.9 &58.1 &43.6 &71.2 &19.2  &45.9 \\
				SGACNet (Ours)              &0.0 &\textbf{29.1} &\textbf{75.5} &36.9 &56.4 &\textbf{56.9} &27.4 &36.5 &\textbf{23.9} &32.2 &\textbf{63.0} &43.2 &21.1 &77.4 &66.4 &47.4 &\textbf{71.7} &18.3 &\textbf{47.8} \\
				\bottomrule
		\end{tabular}}
	\end{center}
\end{table*}

\begin{figure*}[!t]
	\centering
	\subfigure[RGB Inputs]{
		\includegraphics[width=0.945in]{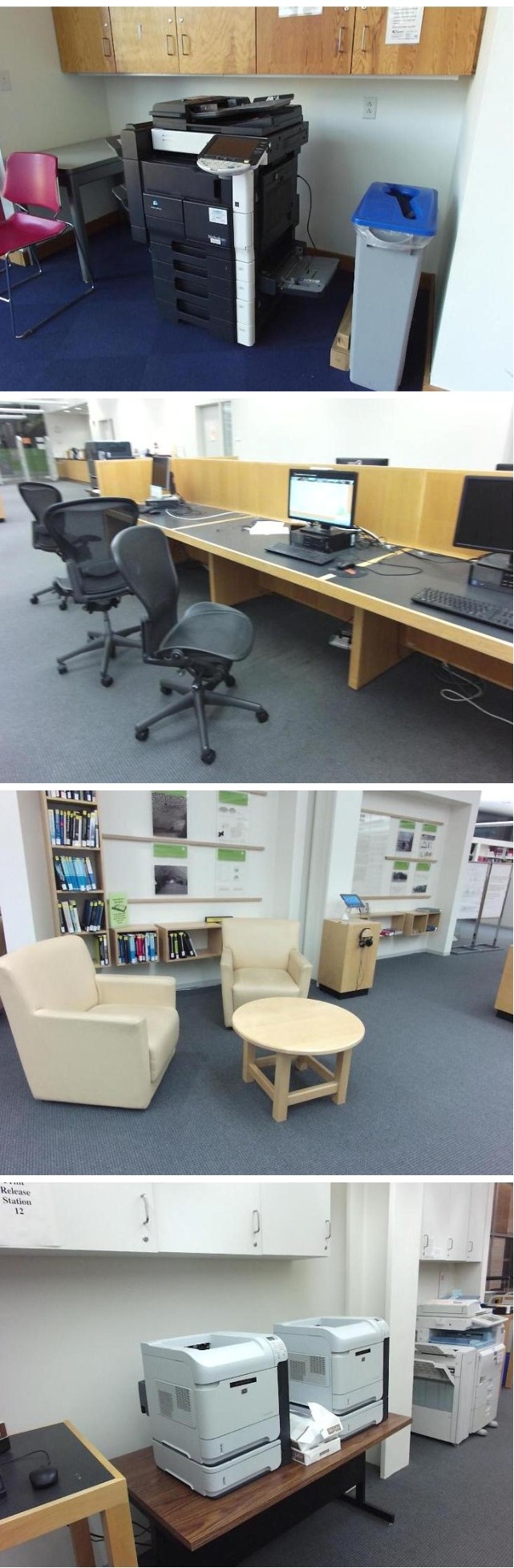}
	}\hspace{-0.5em}
	\subfigure[Depth Inputs]{
		\includegraphics[width=0.945in]{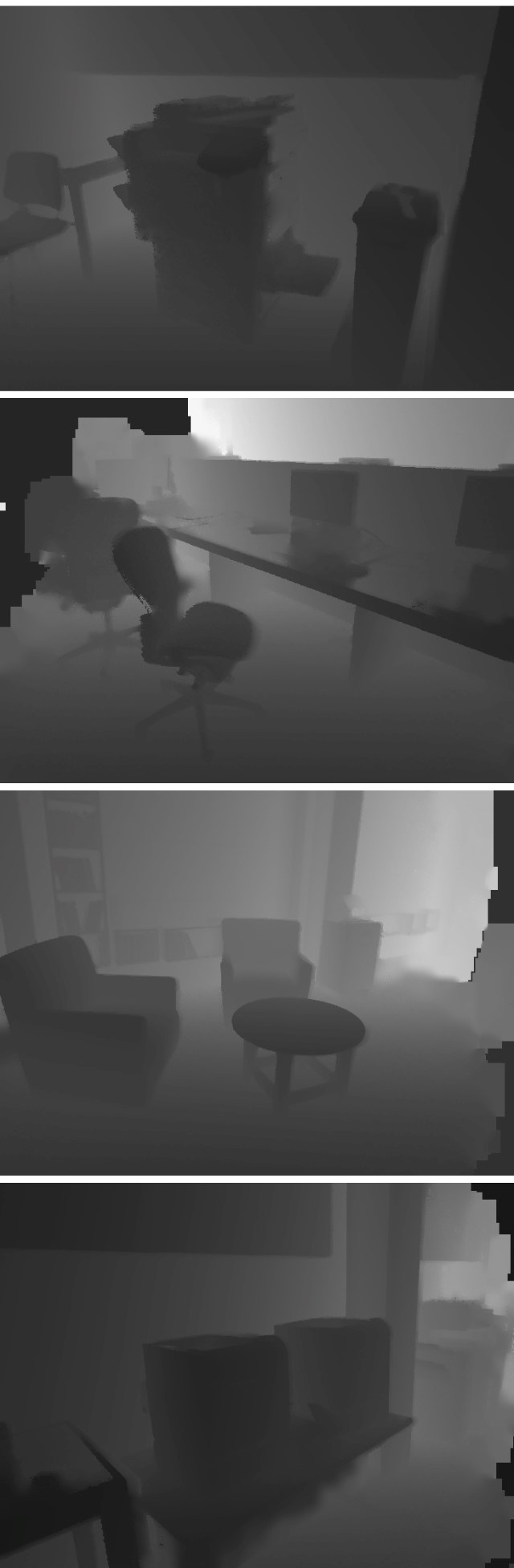}
	}\hspace{-0.5em}
	\subfigure[Ground Truth]{
		\includegraphics[width=0.945in]{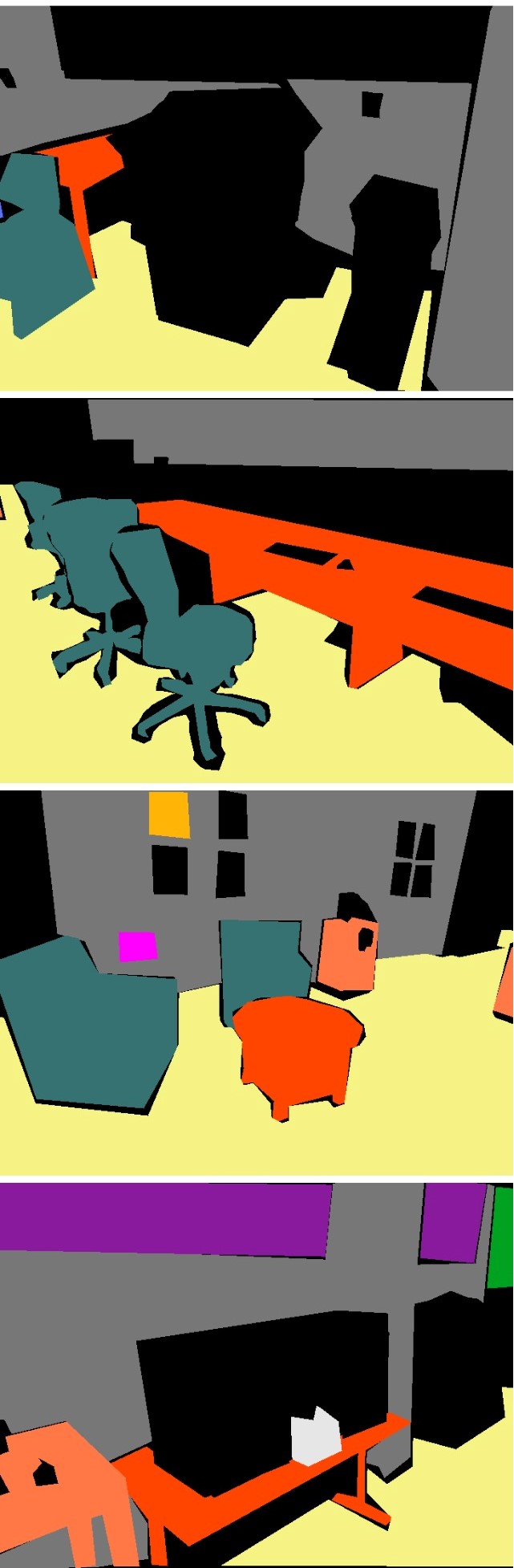}
	}\hspace{-0.5em}
	\subfigure[Baseline-R18]{
		\includegraphics[width=0.945in]{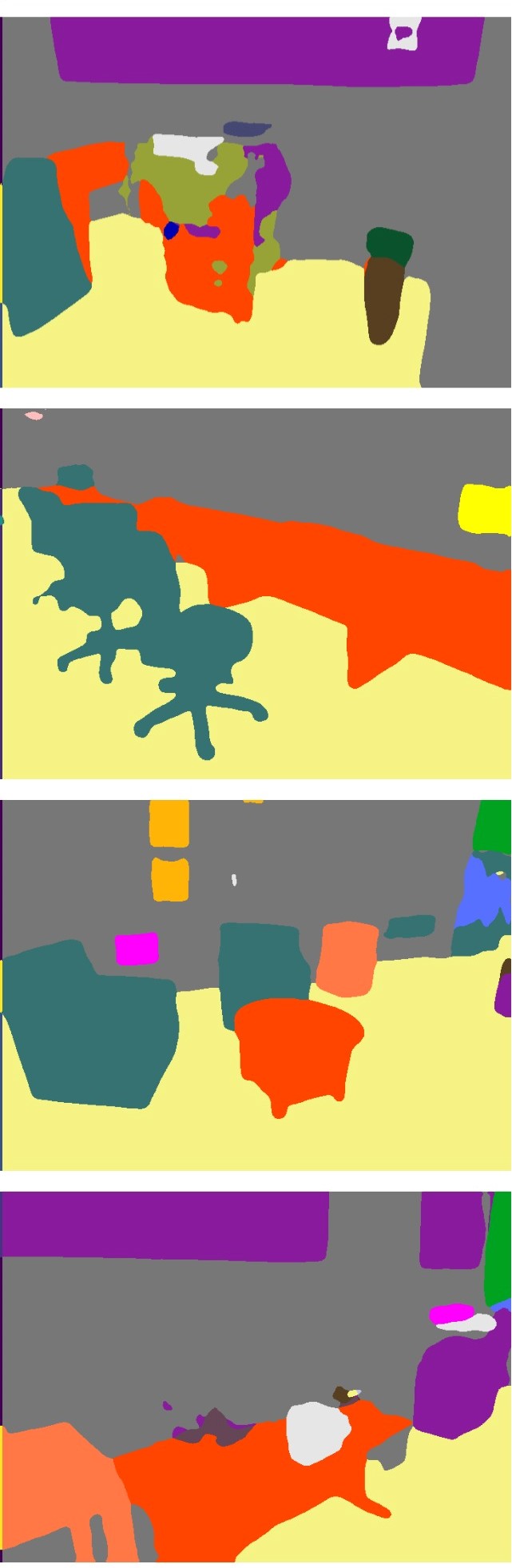}
	}\hspace{-0.5em}
	\subfigure[Baseline-R34]{
		\includegraphics[width=0.945in]{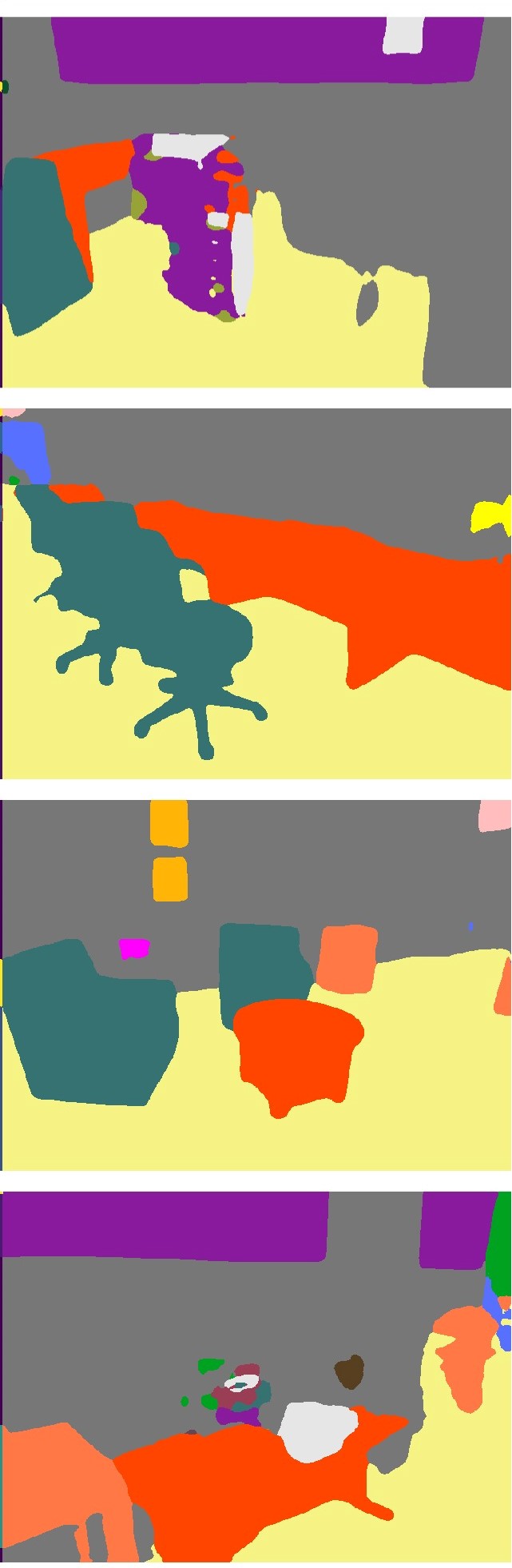}
	}\hspace{-0.5em}
\subfigure[Ours-R18]{
	\includegraphics[width=0.945in]{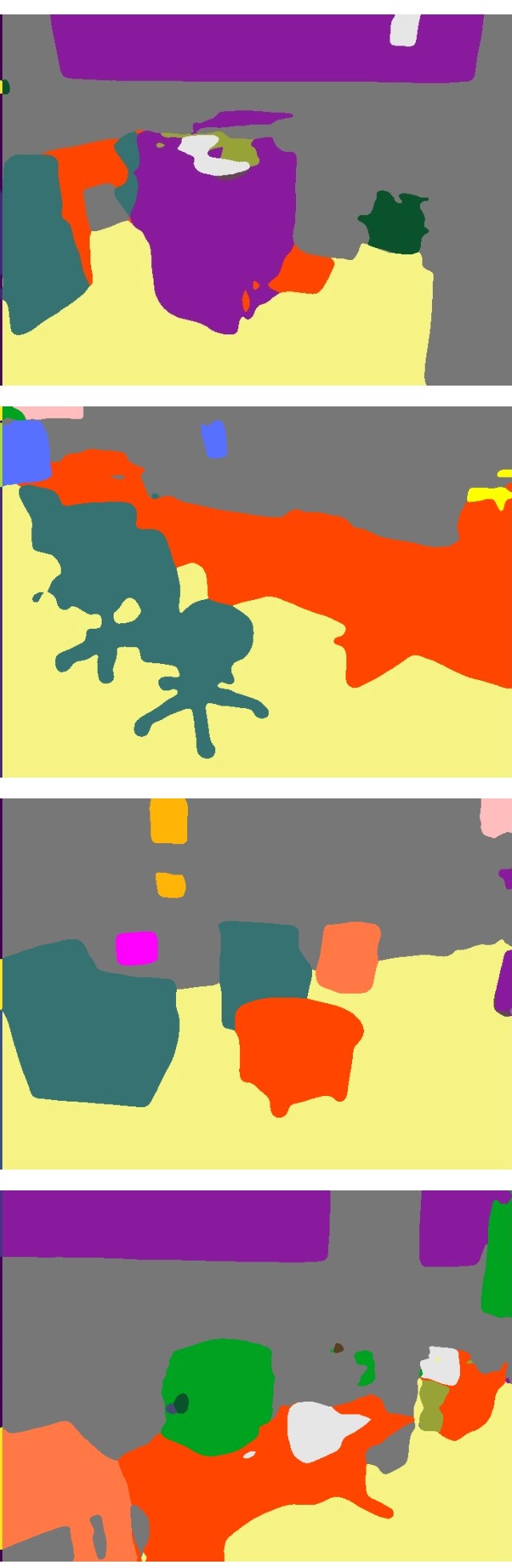}
}\hspace{-0.5em}
	\subfigure[Ours-R34]{
		\includegraphics[width=0.945in]{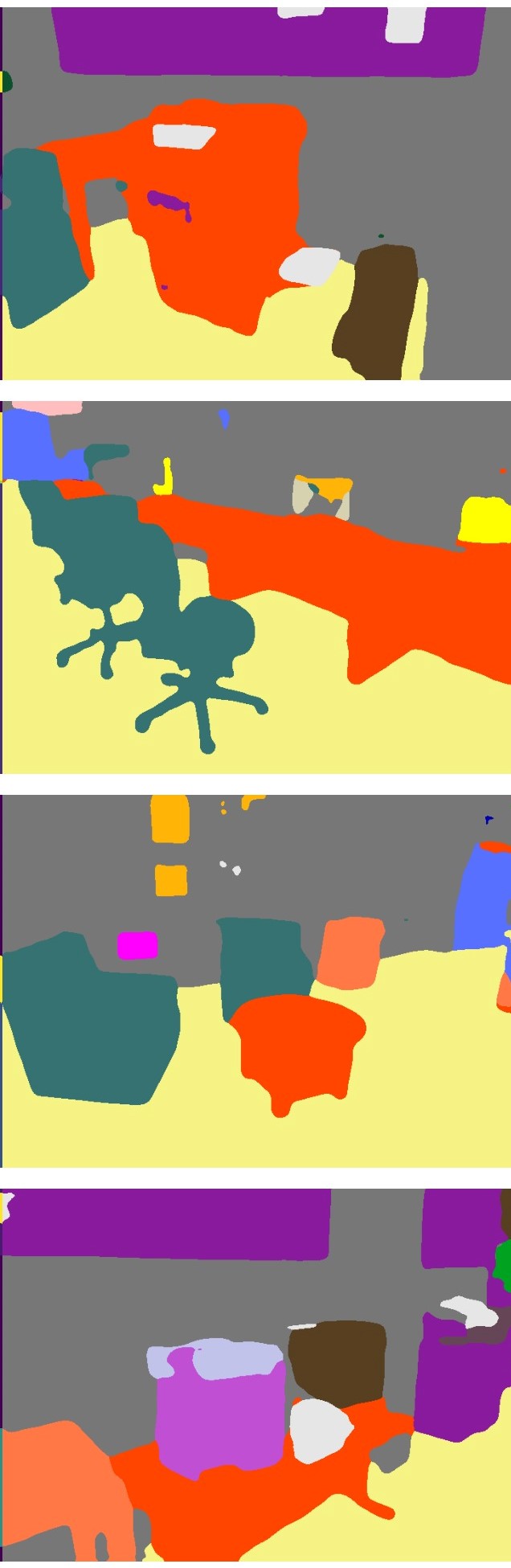}
	}\hspace{-0.5em}
	\caption{Qualitative comparison results of semantic segmentation on SUN RGB-D dataset.  }
	\label{intro}
  \end{figure*}

\begin{table}[!t]
	\centering
	\renewcommand{\arraystretch}{1.08}
		\caption{
			Comparison results with the state-of-the-art approaches on NYUv2 datasets. SI: segmentation input. *: reproduced by our experiment.
		}
		\label{tab:sota}
		\footnotesize
        \setlength{\tabcolsep}{1.6mm}{
		\begin{tabular}{llccccccc}
			\toprule
			\multicolumn{1}{l}{\textbf{Methods}}     &
			\multicolumn{1}{l}{\textbf{Backbone}}     &
			\multicolumn{1}{c}{\textbf{SI}}     &
			\multicolumn{1}{c}{\textbf{mIoU}}            &
			\multicolumn{1}{c}{\textbf{PixAcc.} }            &
			\multicolumn{1}{c}{\textbf{mAcc.} }             
			\\ \midrule
			
             DRD~\cite{DRD}          		&VGG-16 	    &  RGB   &  38.2 & 51.0 & --   \\
             MTI-Net~\cite{MTI-Net}  		&HRNet48     &  RGB  & 49.0 & 75.3 &  62.9\\
           RefineNet~\cite{RefineNet}&ResNet152     &  RGB  & 47.6 & 74.4 &  59.6  \\
            FDNet~\cite{FDNet}& DenseNet264     &  RGB  &  47.4 & 73.9 & 60.3   \\
			\midrule

           RAFNet~\cite{RAFNet}    		&ResNet50		& RGB-D & 47.5 &  73.8 & 60.3        \\
           ACNet~\cite{ACNet}             	&ResNet50		& RGB-D & 48.3 &  --   &  \textbf{63.1}    \\
           GED~\cite{GED}             	&ResNet101		& RGB-D & 48.4 &  75.1 &  61.2 \\
           3DGNN~\cite{3DGNN}      		&ResNet101	    &  RGB-D  & 48.6 & 75.0 &  61.7     \\
           SGNet~\cite{SGNet}             	&ResNet101		& RGB-D & 49.0 &  \textbf{76.4} &  62.7  \\
        MCN-DRM\cite{MCN-DRM}               &MCN-DRM      	& RGB-D & 43.1 &  -- &  56.1    \\
        ESANet*~\cite{ESANet}           &R34-NBt1D		& RGB-D & 49.1 &  75.7 &  62.2 \\		
           TSNet~\cite{TSNet}      	&R34/VGG-16      	& RGB-D & 46.1 &  73.5 &  59.6  \\
           \midrule
			SGACNet (Ours)                  &R18-NBt1D		& RGB-D & 48.2 &  74.6 &  61.8  \\
			SGACNet (Ours)                  &R34-NBt1D		& RGB-D & \textbf{49.4} &  75.6 &  62.7   \\
			\bottomrule
		\end{tabular}}%
\end{table}%

\begin{table}[!t]
	\centering
	\renewcommand{\arraystretch}{1.08}
		\caption{
			Comparison results with the state-of-the-art approaches on SUN RGB-D datasets. SI: segmentation input. *: reproduced by our experiment.
		}
		\label{tab:sota1}
		\footnotesize
        \setlength{\tabcolsep}{1.6mm}{
		\begin{tabular}{llccccccc}
			\toprule
			\multicolumn{1}{l}{\textbf{Methods}}     &
			\multicolumn{1}{l}{\textbf{Backbone}}     &
			\multicolumn{1}{c}{\textbf{SI}}    &
			\multicolumn{1}{c}{\textbf{mIoU}}            &
			\multicolumn{1}{c}{\textbf{PixAcc.}}             &
			\multicolumn{1}{c}{\textbf{mAcc.}}             
			\\ \midrule
             DRD~\cite{DRD}          		&VGG-16 	    &  RGB   &  39.5 & 48.9 & --   \\
               EFCN-8s~\cite{EFCN}&VGG-16     &  RGB  &  40.7 & 76.9 & 53.5   \\
			RefineNet~\cite{RefineNet}      		&ResNet152	    &  RGB  &  47.0 & 81.1 &  57.7    \\
            DSNet~\cite{DSNet}& MobileNetV2     &  RGB  &  32.1 & 75.6 & --  \\
			\midrule
           RAFNet~\cite{RAFNet}            &ResNet50 	 	& RGB-D & 47.2  & 81.3 & 59.4   \\
            CMANet~\cite{CMANet}           &ResNet50		& RGB-D & 47.2  & 81.1 & 59.3   \\
			ACNet~\cite{ACNet}             	&ResNet50		& RGB-D &\textbf{48.1} &  --  & 60.3   \\
           3DGNN~\cite{3DGNN}          		&ResNet101 	    &  RGB-D   &  45.9 &81.8 & 57.5   \\
			SGNet~\cite{SGNet}             	&ResNet101		& RGB-D & 47.1  & 81.0 & 59.6   \\
           ShapeConv~\cite{ShapeConv}           &ResNet101		& RGB-D & 47.6  & \textbf{82.0} & 58.5   \\
			 MCN-DRM~\cite{MCN-DRM}            &MCN-DRM  	& RGB-D & 42.8  & -- & 54.6   \\
            ESANet*~\cite{ESANet}           &R34-NBt1D		& RGB-D & 47.5  & 81.6 & 58.8   \\
            \midrule
			SGACNet (Ours)                  &R18-NBt1D		& RGB-D & 46.5  &81.0  & 57.8   \\
			SGACNet (Ours)                  &R34-NBt1D		& RGB-D & 47.8  &81.2  &\textbf{60.8}  \\
			\bottomrule
		\end{tabular}}%
\end{table}%

\subsection{Ablation Study}\label{sec:Ablation}
We further explore the influence of each component of our proposed segmentation model on NYUv2 dataset. The ablation experimental results are summarized in Table~\ref{tab:AFM}. 

\subsubsection{Attention Fusion Module} In this section, we primarily take ResNet18 as the backbone of our experimental network and then apply several positive attention mechanisms to obtain intuitive visual effects in Figs.~\ref{AFM}(c)-(g).
Performance of dual SE~\cite{SENet} is excellent for extracting and enhancing object boundary information. It can be seen from Fig.~\ref{AFM}(c) that the color of the edge is more obvious, and in the middle area of the object, since there is usually a large amount of similar background information, it is more difficult for the model to determine which channels are useful, so there are fewer pixels in these areas. On the contrary, dual SPGE~\cite{SpatialGE} and dual SPA147~\cite{spa} are superior in expressing the relationships between different regions from spatial position aspects. 
Based on the above analysis, it is unsurprising that the combination of channel and spatial information can achieve better results. As SPGE uses group convolution, which has a smaller receptive field and is less capable of capturing more global features in the center of the object. Therefore, SE+SPA can better handle features between the interior and boundaries of objects and produce clearer attention maps.

\subsubsection{Context and Decoder Module}
We conduct ablation experiments for the context and decoder parts, respectively. The second part of Table~\ref{tab:AFM} shows that APC can improve the segmentation performance while slightly reducing the inference speed. When replacing the light-weighted decoder with a normal decoder module, it obtains 47.9\% mIoU, which is the same as the baseline(first data line). But under this design, LD can significantly decrease the number of computing parameters by about 33.7$\%$. Compared with L3$\times$3, bilinear upsampling can typically achieve better image reconstruction and higher accuracy, reaching 48.2\% mIoU.

\subsection{Comparison with SOTA Methods}
\subsubsection{Comparison with SOTA on each category}
In Table~\ref{tab:imbalance1} and Table~\ref{tab:imbalance2}, we provide segmentation results for each class. Since we not only focus on metrics like accuracy but also on imbalanced distributed data. Surprisingly, we have to admit that the implementation of our method on some categories (\textit{e.g.} wall, cabinet, dresser) is satisfactory. It may be that the noise contained in the depth map affects the accurate contour representation, and the RGB map is confused by lighting conditions as shown in Figs.~\ref{AFM}(a)-(b). At the same time, our SGACNet is beneficial from AFM, which could extract information from RGB and depth maps, resulting in balanced fused features for segmentation. 

Furthermore, we also observed that all methods in Table \ref{tab:imbalance2} have low accuracy in floormat. We speculate that this could be due to the lack of dedicated annotated data for floormat in the SUN RGB-D, which makes it challenging for algorithms to effectively learn and generalize to this specific category. Finally, our SGACNet has achieved the highest IoU values in 18 out of 40 classes in Table~\ref{tab:imbalance1} and the best performance in 16 out of 37 classes in Table~\ref{tab:imbalance2}. These experimental results indicate that our SGACNet is more reliable in terms of class-balanced segmentation performance compared to existing SOTA methods.

\subsubsection{Comparison results on NYUv2}
As shown in Table~\ref{tab:sota}, SGACNet gains the best mIoU of 49.4$\%$, and our experimental results with three segmentation metrics (mIoU, PixAcc., mAcc.) are comparable to those using deeper networks such as GED \cite{GED}, 3DGNN \cite{3DGNN}, SGNet \cite{SGNet}. These results demonstrate that the design of SGACNet is both reasonable and efficient. Moreover, it can be seen that if we take good care of depth modality different from the RGB method, we will usually get more essential indicators for model evaluation. 

\subsubsection{Comparison results on SUN RGB-D}
In Table~\ref{tab:sota1}, ACNet~\cite{ACNet}, ShapeConv~\cite{ShapeConv}, and CMANet~\cite{CMANet} prove that a well-designed RGB-D segmentation method can achieve favorable overall performance, despite the inevitable increase in computational complexity. In contrast to models that do not handle image segmentation efficiently, our SGACNet achieves a balance of three accuracy metrics, with 47.8$\%$ mIoU, 81.2$\%$ PixAcc., and 60.8$\%$ mAcc., which is close to the performance of the baseline (Params of 46.95M)~\cite{ESANet} with fewer parameters (Params of 35.64M).

\subsubsection{Comparison results on Cityscapes}
To fully evaluate the performance of SGACNet, we conducted comparative experiments on the Cityscapes dataset, which is commonly used in autonomous driving and other fields. As shown in Table~\ref{tab:city}, when the input resolution is 512$\times$1024 pixels, SGACNet achieves segmentation accuracies that are 10.7$\%$ and 7.6$\%$ higher than those of ESPNet-v2~\cite{ESPNetv2} and LDFNet~\cite{LDFNet}, respectively. Additionally, it achieves similar performance to ESANet~\cite{ESANet}, while using 24.09$\%$ fewer parameters. However, when the input resolution is 1024$\times$2048 pixels, compared with SwiftNet~\cite{SwiftNet}, ICNet~\cite{ICNet}, our method is difficult to achieve the corresponding effect, indicating that RGB input cloud is better suited for real-time semantic segmentation tasks with high-resolution inputs. 

\subsection{Inference Speed}
In addition to extensive experiments in accuracy and model parameters, we then deploy the proposed method on a single NVIDIA RTX 3090Ti for speed inference, including \textbf{all the steps} for semantic segmentation, such as loading data, transforming, inference, etc. As shown in Table~\ref{tab:TIME}, practical results prove that our model is not only suitable for pixel-level semantic segmentation, but it is also capable of capturing some motion detail at an average of 16 FPS in different scenes. Under the condition of the half resolution, our highest inference speed of ResNet18 can reach 20.7 FPS. Therefore, our proposed SGACNet achieves a trade-off among segmentation accuracy, inference time, and parameters.

\begin{table}[!t]
	\centering
	\renewcommand{\arraystretch}{1.08}
	\caption{Comparison results of semantic segmentation with the state-of-the-art methods on Cityscapes dataset which includes full resolution (1024$\times$2048) and the half resolution (512$\times$1024). SI: segmentation input. *: reproduced by our experiment.}
	\footnotesize
	\begin{tabular}{llcccc}
		\toprule
		\multicolumn{1}{l}{\multirow{2}{*}{\textbf{Methods}}}     &
		\multicolumn{1}{l}{\multirow{2}{*}{\textbf{Backbone}}}     &
		\multicolumn{1}{c}{\multirow{2}{*}{\textbf{SI}}}     &
		\multicolumn{1}{c}{\multirow{2}{*}{\textbf{Params}$\downarrow$}} 						&
		\multicolumn{2}{c}{\textbf{mIoU$\uparrow$}} \\ 
		\cmidrule(r){5-6} 
		\multicolumn{4}{c}{}                                  &      
		\multicolumn{1}{c}{\textbf{half}}      & 
		\multicolumn{1}{c}{\textbf{full}}	     \\ 
		\midrule
		Lite-HRNet~\cite{Lite-HRNet}       & HRNet18          & RGB   & --  &  73.8 &  --       \\
		SwiftNet~\cite{SwiftNet}           & ResNet18   	& RGB   & \textbf{11.1}&  70.2 & 75.4     \\
      ESPNet-v2~\cite{ESPNetv2}          & ESPNetv2   	& RGB   & --  &  66.2 &  --       \\
      ESANet*~\cite{ESANet}              & ResNet34      & RGB   & 32.1&  72.5 & 78.7    \\
		ICNet~\cite{ICNet}                 & PSPNet50  	& RGB   & 26.5 &   --  & 69.5       \\
		SGACNet(Ours)                      & ResNet18  	& RGB   & 14.0&  71.2 &77.5 \\
		SGACNet(Ours)                      & ResNet34  	& RGB   & 20.8&  72.3 & 78.5    \\ 
		\midrule
		LDFNet~\cite{LDFNet}               & ERFNet 	    & RGB-D &  -- &  68.5 & --        \\
        ESANet*~\cite{ESANet}              & R34-NBt1D      & RGB-D & 46.9&  \textbf{74.6} & \textbf{79.8}    \\
        \midrule
		SGACNet(Ours)                      & R18-NBt1D      & RGB-D & 22.1&  73.3 & 78.7 \\
		SGACNet(Ours)                      & R34-NBt1D     & RGB-D & 35.6&  74.1 & 79.7  \\
		\bottomrule
	\end{tabular}%
	\label{tab:city}%
\end{table}%

\begin{table}[!t]%
	\centering
	\renewcommand{\arraystretch}{1.08}
	\caption{Comparison results of inference speed (FPS) with baseline method on three datasets including \textbf{all the steps} for semantic segmentation. We take ResNet18 and ResNet34 as our backbones. SI: segmentation input.}
	\footnotesize
	\begin{tabular}{llccc}
		\toprule
		\textbf{Datasets} & \textbf{SI}     &\textbf{ResNet18}   & \textbf{ResNet34}            \\
		\midrule
		NYUv2 		& RGB-D & 15.9 & 13.8 \\ 
		SUNRGB-D 	& RGB-D & 16.7 & 14.3 \\ 
		Cityscapes.full 	& RGB   & 16.7 & 18.6 \\
		Cityscapes.half 	& RGB   & 20.7 & 18.9 \\ 
		Cityscapes.full     & RGB-D & 17.3 & 16.4  \\
		Cityscapes.half     & RGB-D & 17.6 & 16.5  \\ 
		\bottomrule
	\end{tabular}%
	\label{tab:TIME}%
\end{table}%

\section{Conclusions}\label{sec:Conclusions}
In this paper, we have designed an efficient light-weighted network for RGB-D semantic segmentation called SGACNet. This model can obtain a larger receptive field and richer semantic information in dual different attention fusion and contextual information learning. In addition, we incorporate a lightweight decoder to reduce computation, thus achieving a trade-off among segmentation accuracy, inference speed, and model parameters. Experiments show that our proposed SGACNet has the ability of efficient semantic segmentation, which can achieve comparable performance with nearly 40.1\% model parameter reduction compared with the state-of-the-art models on NYUv2, SUN RGB-D, and Cityscapes datasets. In the future, we will further explore the ability of the channel-space fusion attention module to extract depth information and also consider deployment in the NVIDIA Jetson series or other edge devices. We also plan to achieve video semantic segmentation with low memory usage and timing correlation.

\bibliographystyle{IEEEtran}
\bibliography{cvmbib}

\end{document}